\documentclass[times, review, 10pt]{elsarticle}

\usepackage{microtype}
\usepackage{graphicx}
\usepackage{caption}
\usepackage{subcaption}
\usepackage{booktabs} 
\usepackage{wrapfig}
\usepackage{pifont}
\usepackage{placeins}
\usepackage{soul}
\usepackage{enumitem}
\usepackage{bbm}
\usepackage{color,soul}
\usepackage{multirow}

\usepackage{hyperref}

\usepackage{amsmath}
\usepackage{amssymb}
\usepackage{mathtools}
\usepackage{amsthm}
\usepackage{float}
\usepackage{algorithm}
\usepackage{algorithmic}

\newcommand{\algcomment}[1]{\hfill \textcolor{black!60}{// #1}}

\usepackage[capitalize]{cleveref}

\crefformat{equation}{Eq.~\textcolor{blue}{(#2#1#3)}}
\Crefformat{equation}{Eq.~\textcolor{blue}{(#2#1#3)}}
\crefrangeformat{equation}{Eqs.~\textcolor{blue}{(#3#1#4)}--\textcolor{blue}{(#5#2#6)}}
\Crefrangeformat{equation}{Eqs.~\textcolor{blue}{(#3#1#4)}--\textcolor{blue}{(#5#2#6)}}
\crefmultiformat{equation}{Eqs.~\textcolor{blue}{(#2#1#3)}}{ and~\textcolor{blue}{(#2#1#3)}}{, \textcolor{blue}{(#2#1#3)}}{, and~\textcolor{blue}{(#2#1#3)}}

\crefformat{figure}{Fig.~\textcolor{blue}{#2#1#3}}
\Crefformat{figure}{Fig.~\textcolor{blue}{#2#1#3}}
\crefrangeformat{figure}{Figs.~\textcolor{blue}{#3#1#4}--\textcolor{blue}{#5#2#6}}
\Crefrangeformat{figure}{Figs.~\textcolor{blue}{#3#1#4}--\textcolor{blue}{#5#2#6}}

\crefmultiformat{figure}{Figs.~\textcolor{blue}{#2#1#3}}{ and~\textcolor{blue}{#2#1#3}}{, \textcolor{blue}{#2#1#3}}{, and~\textcolor{blue}{#2#1#3}}

\crefformat{table}{Table~\textcolor{blue}{#2#1#3}}
\Crefformat{table}{Table~\textcolor{blue}{#2#1#3}}
\crefrangeformat{table}{Tables~\textcolor{blue}{#3#1#4}--\textcolor{blue}{#5#2#6}}
\Crefrangeformat{table}{Tables~\textcolor{blue}{#3#1#4}--\textcolor{blue}{#5#2#6}}
\crefmultiformat{table}{Tables~\textcolor{blue}{#2#1#3}}{ and~\textcolor{blue}{#2#1#3}}{, \textcolor{blue}{#2#1#3}}{, and~\textcolor{blue}{#2#1#3}}

\hypersetup{
    colorlinks=true,
    linkcolor=blue,  
    citecolor=blue,  
    urlcolor=blue    
}

\theoremstyle{plain}
\newtheorem{theorem}{Theorem}[section]

\newtheorem{lemma}[theorem]{Lemma}
\newtheorem{corollary}[theorem]{Corollary}
\theoremstyle{definition}
\newtheorem{definition}[theorem]{Definition}
\newtheorem{assumption}[theorem]{Assumption}
\theoremstyle{remark}
\newtheorem{remark}{Remark}[section]

\usepackage[textsize=tiny]{todonotes}
\providecommand{\mathbbm}[1]{\mathbb{#1}}

\makeatletter
\renewcommand\appendix{\par
  
  \setcounter{section}{0}%
  \setcounter{subsection}{0}%

  \gdef\thesection{Appendix}%
  \gdef\thesubsection{\@Alph\c@section.\arabic{subsection}}%
  \gdef\thefigure{\arabic{figure}}%
  \gdef\thetable{\arabic{table}}%

  \renewcommand{\thetheorem}{\Alph{section}.\arabic{theorem}}%
  \renewcommand{\thelemma}{\Alph{section}.\arabic{lemma}}%
  \renewcommand{\thecorollary}{\Alph{section}.\arabic{corollary}}%
  \renewcommand{\thedefinition}{\Alph{section}.\arabic{definition}}%
  
  \renewcommand{\theassumption}{\Alph{section}.\arabic{assumption}}%
  \renewcommand{\theproposition}{\Alph{section}.\arabic{proposition}}%
  \renewcommand{\theremark}{\Alph{section}.\arabic{remark}}%
}
\makeatother

\begin{document}

\begin{frontmatter}

\title{Communication-Efficient Neural Tangent Kernels for Heterogeneous Decentralized Federated Learning}

\author[inst1]{Li Xia\corref{cor1}}
\ead{xiali@muc.edu.cn}

\cortext[cor1]{Corresponding author}

\address[inst1]{organization={Key Laboratory of Ethnic Language Intelligent Analysis and Security Governance of MOE, Minzu University of China},
            city={Beijing},
            country={China}}

\begin{abstract}
Decentralized federated learning (DFL) enables collaborative model training without a central server, but converges slowly under statistical heterogeneity. Recent work has shown that neural tangent kernel (NTK) methods achieve faster convergence than gradient-based updates in DFL, while momentum has proven effective for accelerating gradient-based FL. However, applying momentum to NTK updates can destabilize training under heterogeneous data. We propose SPARK, which addresses this instability with a stage-wise annealed soft-label regularizer evaluated on neighborhood-aggregated data, so that momentum can accelerate NTK updates stably. Under high heterogeneity, SPARK converges about 3$\times$ faster than baselines and lowers the total communication to a target accuracy by up to about 70\%, and it attains higher accuracy across heterogeneity levels. We further study random projection as an optional Jacobian-compression strategy for bandwidth-constrained settings. We validate the approach across multiple datasets, network topologies, and heterogeneity levels. 
\end{abstract}

\begin{keyword}
Decentralized Federated Learning \sep Neural Tangent Kernel \sep Heterogeneous distributions \sep Knowledge Distillation 
\end{keyword}

\end{frontmatter}


\section{Introduction}
\label{sec:introduction}

The proliferation of Internet of Things (IoT) devices such as industrial sensors, mobile terminals, and wearable monitors generates large volumes of data with diverse local distributions~\cite{li2025robust}. This statistical heterogeneity complicates collaborative learning, since local objectives differ across devices~\cite{guo2024dynamic,kang2024fednn}. Federated learning (FL)~\cite{fedavg,li2020federated} enables collaborative training without sharing raw data, but its reliance on a central server introduces communication bottlenecks, single points of failure, and limited scalability~\cite{fed_open_problems}. Decentralized federated learning (DFL)~\cite{DFL_survey,dfl_survey_2} removes the central server by letting devices communicate over a peer-to-peer graph, which makes it well suited to distributed IoT deployments.

While DFL alleviates the drawbacks of centralized FL, it still grapples with statistical heterogeneity across devices, which slows convergence. Traditional first-order gradient-based methods~\cite{d-psgd,dfedavg,dfedsam,dispfl} converge slowly on non-IID data and often suffer from device drift, where local updates diverge from the global optimum. Recent NTK-based methods~\cite{ntkdfl} instead evolve weights through Jacobian matrices and converge faster under heterogeneity. However, transmitting full Jacobian matrices incurs much higher communication than transmitting weights, which limits their use in bandwidth-constrained environments. Standard compression techniques designed for first-order gradients~\cite{alistarh2017qsgd,lin2017deep,vogels2019powersgd} disrupt the spectral structure of the Jacobian on which kernel construction depends. While momentum acceleration is effective in distributed optimization~\cite{polyak1964momentum,nesterov1983method,dfedavg}, combining it with NTK updates under heterogeneity remains difficult because of noise-induced instability.

We propose SPARK (Stage-wise Projected NTK and Accelerated Regularization), a framework that enables stable momentum acceleration of NTK updates under heterogeneity. Its core components are stage-wise annealed distillation and momentum acceleration, complemented by random projection as a communication-reduction strategy. To stabilize momentum-accelerated updates, stage-wise annealed distillation gradually introduces soft-label regularization computed on the aggregated neighborhood data. Momentum is then applied to the stabilized updates to accelerate convergence. To further reduce communication in bandwidth-constrained settings, SPARK can optionally compress Jacobians via random projection while approximately preserving their spectral structure. This design reduces communication while preserving the convergence benefits of NTK-based updates. The contributions of this work are threefold:
\begin{enumerate}[left=0pt]
    \item We propose SPARK, a decentralized federated learning framework that, to the best of our knowledge, is the first to safely combine momentum with NTK-based weight evolution under non-IID data, using stage-wise annealed distillation to stabilize the accelerated updates.
    
    \item SPARK converges about 3$\times$ faster than NTK-based baselines to 85\% accuracy and attains higher final accuracy, which reduces the total communication needed to reach a target accuracy through fewer rounds.
    
    \item We provide a non-convex convergence analysis for SPARK and show consistent gains across datasets, network topologies, and heterogeneity levels, and we further study random projection as an optional Jacobian-compression strategy for bandwidth-constrained settings.
\end{enumerate}


\section{Related Work}

\subsection{Federated and decentralized learning}
\label{subsec:fl_dfl}
Federated learning (FL), introduced by McMahan et al.~\cite{fedavg}, enables collaborative model training across devices without sharing raw data, addressing privacy and compliance concerns~\cite{zhu2023surrogate,xia2025nonlinear}. Recent studies have addressed statistical heterogeneity in FL through regularization and dynamic aggregation~\cite{karimireddy2020scaffold,li2020federated,guo2024dynamic}. However, the central-server architecture introduces communication bottlenecks and single points of failure~\cite{fed_open_problems}. Decentralized federated learning (DFL) removes the central server by connecting clients in peer-to-peer topologies~\cite{DFL_survey}. Representative DFL methods include D-PSGD~\cite{d-psgd}, DFedAvg~\cite{dfedavg}, and DFedSAM~\cite{dfedsam}. Li et al.~\cite{li2025robust} further study robust DFL under non-ideal communication, while Mao et al.~\cite{mao2025fedkt} use knowledge transfer to handle non-IID data in centralized FL. These methods are mainly built on first-order model or gradient updates, leaving the use of higher-order NTK dynamics under decentralized communication less explored.

\subsection{Neural tangent kernel in distributed learning}
\label{subsec:ntk_distributed}
The neural tangent kernel framework~\cite{ntk_jacot} provides a principled theoretical foundation for analyzing infinite-width neural networks through kernel-based linearization~\cite{ntk_linearity}. Recent work has applied NTK theory to federated settings. Huang et al.~\cite{huang2021fl} analyzed FedAvg convergence through the NTK lens, while Yu et al.~\cite{yu2022tct} demonstrated that convex networks can be trained using NTK in centralized FL. Building on these foundations, Yue et al.~\cite{yue2022ntk} replaced SGD with NTK-based weight evolution. Most notably, Thompson et al.~\cite{ntkdfl} extended NTK to decentralized environments, showing faster convergence than existing DFL methods under heterogeneous data distributions. While NTK-DFL improves convergence through expressive kernel-based updates, it incurs higher communication cost, because transmitting full Jacobian matrices requires more bandwidth than weight-based methods~\cite{comm_efficient}. This leaves the Jacobian communication overhead of NTK-based DFL largely unresolved in bandwidth-constrained edge environments.

\subsection{Communication efficiency and acceleration}
\label{subsec:comm_efficiency}
A large body of work improves communication efficiency through gradient 
quantization~\cite{alistarh2017qsgd}, sparsification~\cite{lin2017deep}, and 
low-rank factorization~\cite{vogels2019powersgd}. Unlike these
first-order schemes, NTK-based DFL communicates Jacobian matrices whose
inner-product geometry must be preserved for kernel construction, so standard
compression does not transfer directly.
Random projection~\cite{johnson1984extensions,li2018measuring} and knowledge 
distillation~\cite{fedagent,guo2025parameterized,mao2025fedkt} have been studied 
separately for dimensionality reduction and knowledge transfer, 
but their combination for communication-efficient NTK-based DFL remains 
largely unexplored. Momentum acceleration yields convergence 
speedups in FL~\cite{cheng2024momentum} and has been applied 
to first-order DFL methods~\cite{dfedavg}. Nevertheless, 
combining momentum acceleration with NTK-based DFL remains 
challenging under heterogeneous data, and Jacobian compression can further 
destabilize momentum-based updates through approximation noise.

\begin{figure*}[t]
    \centering
    \includegraphics[width=0.95\textwidth]{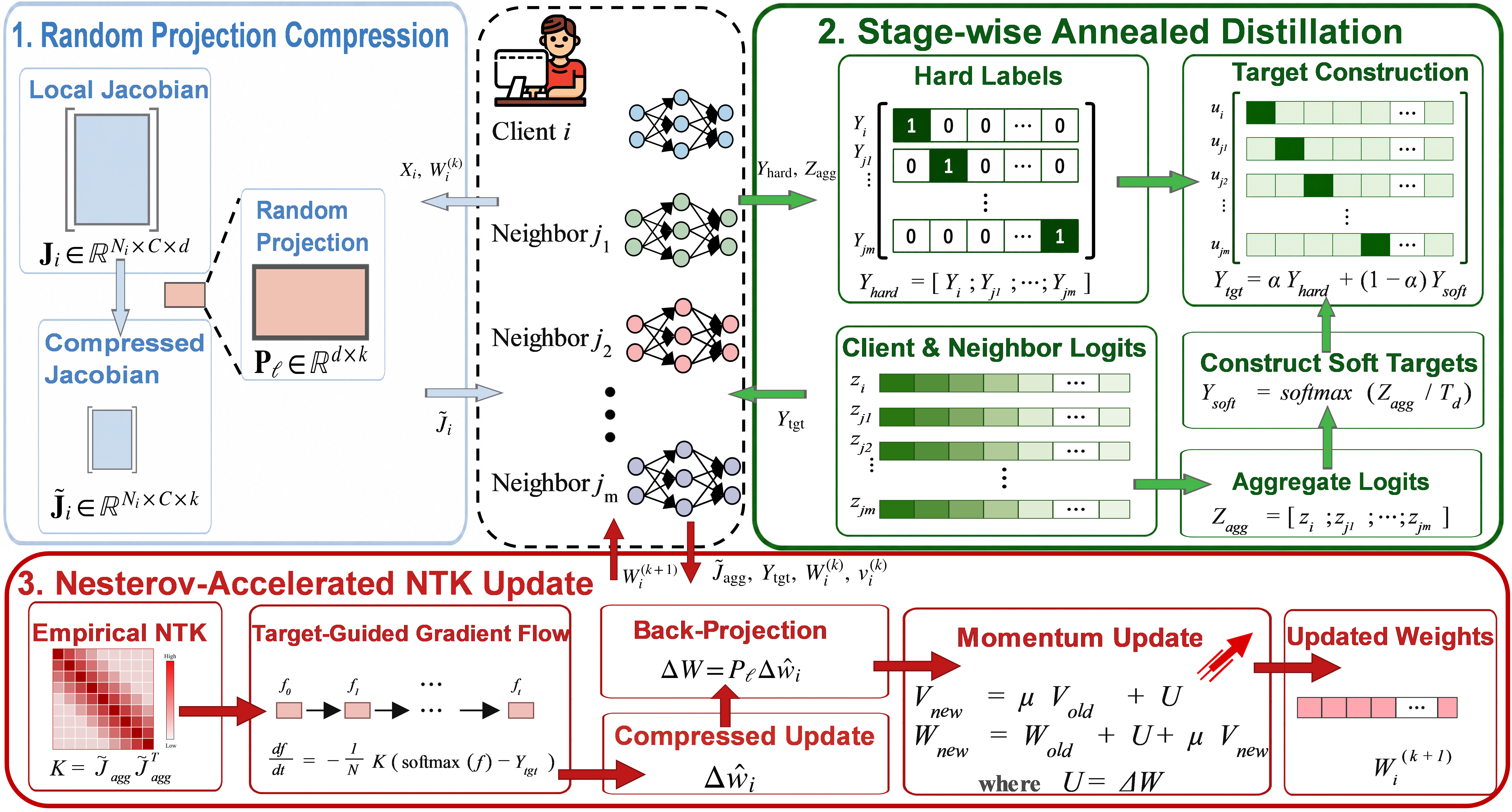}
    \caption{Overview of the SPARK framework: temporal stabilization via annealed distillation and convergence acceleration via momentum, with random projection as an optional communication-compression strategy.}
    \label{fig:spark_overview}
\end{figure*}

\begin{algorithm}[!t]
\caption{SPARK: Stage-wise Projected NTK and Accelerated Regularization}
\label{alg:spark}
\begin{algorithmic}[1]
\REQUIRE Graph $\mathcal{G}$, weights $\{\mathbf{w}_i^{(0)}\}$, optional proj. cap $k_{\mathrm{proj}}$, warm-up $R_{\text{warm}}$, total $R$, momentum $\mu$, hyperparams.
\ENSURE Final client models $\{\mathbf{w}_i^{(R)}\}_{i=1}^M$
\STATE Initialize velocity $\mathbf{v}_i^{(0)} \leftarrow \mathbf{0}$ for all clients $i$
\FOR{round $k = 1, \ldots, R$}
    \STATE $\mathbf{w}_i^{(k)} \leftarrow \textsc{Avg}(\{\mathbf{w}_j^{(k)}\}_{j \in \mathcal{N}_i \cup \{i\}})$; \algcomment{Neighborhood consensus}
    \FOR{client $i \in \{1, \ldots, M\}$ \textbf{in parallel}}
        \STATE $\mathbf{J}_i^{(k)} \leftarrow \textsc{Jacobian}(\mathcal{D}_i; \mathbf{w}_i^{(k)})$; \algcomment{Jacobian at consensus weight}
        \STATE $\tilde{\mathbf{J}}_i^{(k)} \leftarrow \mathbf{J}_i^{(k)} \mathbf{P}$; \algcomment{Use $\mathbf{P}=\mathbf{I}$ unless projection is enabled}
        \STATE $\mathbf{z}_i^{(k)} \leftarrow f(\mathbf{X}_i; \mathbf{w}_i^{(k)})$; \algcomment{Logits from consensus model}
        \STATE \textbf{send} $\{\tilde{\mathbf{J}}_i^{(k)}, \mathbf{z}_i^{(k)}\}$ \textbf{to} $\mathcal{N}_i$; \algcomment{Broadcast to neighbors}
    \ENDFOR
    
    \FOR{client $i \in \{1, \ldots, M\}$ \textbf{in parallel}}
        \STATE Receive $\{\tilde{\mathbf{J}}_j^{(k)}, \mathbf{z}_j^{(k)}\}_{j \in \mathcal{N}_i}$;
        \STATE $\tilde{\mathbf{J}}_{\text{agg}}^{(k)} \leftarrow [\tilde{\mathbf{J}}_i^{(k)}, \{\tilde{\mathbf{J}}_j^{(k)}\}_{j \in \mathcal{N}_i}]$; \algcomment{Concatenate neighbor Jacobians}
        \STATE $\mathbf{K}^{(k)} \leftarrow \tilde{\mathbf{J}}_{\text{agg}}^{(k)} (\tilde{\mathbf{J}}_{\text{agg}}^{(k)})^\top$; \algcomment{Compute empirical NTK}
        \STATE $\mathbf{Y}_{\text{tgt}}^{(k)} \leftarrow \textsc{DistillTarget}(k, \dots)$; \algcomment{See Alg.~\ref{alg:distillation}}
        \STATE $\{\mathbf{f}_t^{(k)}\} \leftarrow \textsc{KernelDescent}(\mathbf{K}^{(k)}, \mathbf{Y}_{\text{tgt}}^{(k)})$; \algcomment{Evolve predictions}
        \STATE $\Delta \mathbf{w}_i^{(k)} \leftarrow \textsc{BackProj}(\tilde{\mathbf{J}}_{\text{agg}}^{(k)}, \mathbf{Y}_{\text{tgt}}^{(k)}, \{\mathbf{f}_t^{(k)}\})$; \algcomment{Map to weight space}
        \STATE $\mathbf{v}_i^{(k+1)} \leftarrow \mu \mathbf{v}_i^{(k)} + \Delta \mathbf{w}_i^{(k)}$; \algcomment{Update velocity}
        \STATE $\mathbf{w}_i^{(k+1)} \leftarrow \mathbf{w}_i^{(k)} + \mu \mathbf{v}_i^{(k+1)} + \Delta \mathbf{w}_i^{(k)}$; \algcomment{Nesterov update}
    \ENDFOR
\ENDFOR
\end{algorithmic}
\end{algorithm}

\begin{algorithm}[!t]
\caption{Stage-wise Annealed Distillation Target Construction}
\label{alg:distillation}
\begin{algorithmic}[1]
\REQUIRE Round $k$, thresholds $R_{\text{warm}}, R$, labels $\mathbf{Y}_{\text{hard}}$, logits $\{\mathbf{z}_j\}$, hyperparams.
\ENSURE Distillation target $\mathbf{Y}_{\text{tgt}}$

\IF{$k \leq R_{\text{warm}}$}
    \STATE $\alpha \leftarrow 1.0, \tau \leftarrow 1.0$; \algcomment{Initialize: Pure NTK phase}
\ELSE
    \STATE $p \leftarrow (k - R_{\text{warm}}) / (R - R_{\text{warm}})$; \algcomment{Calculate progress ratio}
    \STATE $\alpha \leftarrow \alpha_{\text{final}} + \frac{1}{2}(\alpha_{\text{init}} - \alpha_{\text{final}})(1 + \cos(\pi p))$; \algcomment{Update mixing coef.}
    \STATE $\tau \leftarrow \tau_{\text{init}} + (\tau_{\text{final}} - \tau_{\text{init}}) p$; \algcomment{Update temperature}
\ENDIF

\STATE Aggregate neighbor logits: $\mathbf{z}_{\text{agg}} = [\mathbf{z}_i; \mathbf{z}_{j_1}; \ldots; \mathbf{z}_{j_{|\mathcal{N}_i|}}]$
\STATE Compute soft labels: $\mathbf{Y}_{\text{soft}} = \text{softmax}(\mathbf{z}_{\text{agg}} / \tau)$
\STATE Construct target: $\mathbf{Y}_{\text{tgt}} = \alpha \mathbf{Y}_{\text{hard}} + (1 - \alpha) \mathbf{Y}_{\text{soft}}$

\STATE \textbf{return} $\mathbf{Y}_{\text{tgt}}$
\end{algorithmic}
\end{algorithm}

\section{Proposed SPARK}
\label{sec:method}

\subsection{Problem Statement}
\label{subsec:problem}

We consider the decentralized federated learning setting where $M$ IoT devices collaboratively train a model without a central server. Each IoT device $i \in \{1, \ldots, M\}$ possesses a local dataset $\mathcal{D}_i = \{(\mathbf{x}_i^{(n)}, \mathbf{y}_i^{(n)})\}_{n=1}^{N_i}$, where $N_i$ is the number of local training samples. The IoT devices are connected via an undirected communication graph $\mathcal{G}=(\mathcal{V},\mathcal{E})$, where each node is an IoT device and each edge is a communication link. Each IoT device $i$ communicates only with its neighbors $\mathcal{N}_i = \{j \in \mathcal{V} : (i,j) \in \mathcal{E}\}$.

Following NTK-DFL~\cite{ntkdfl}, each IoT device maintains a local model with weights $\mathbf{w}_i$, and the global objective minimizes the empirical risk $\mathcal{L}(\mathbf{w}) = \frac{1}{M} \sum_{i=1}^{M} \mathcal{L}_i(\mathbf{w})$, where
\begin{equation}
\label{eq:local_loss}
\mathcal{L}_i(\mathbf{w}) = \frac{1}{N_i} \sum_{n=1}^{N_i} \ell(f(\mathbf{x}_i^{(n)}; \mathbf{w}), \mathbf{y}_i^{(n)})
\end{equation}
is the local empirical loss on IoT device $i$'s dataset.

However, NTK-DFL achieves fast convergence by transmitting full Jacobian matrices $\mathbf{J}_i^{(k)} \in \mathbb{R}^{N_i \times C \times d}$ at each round $k$, where $C$ is the number of output classes and $d$ is the model parameter dimension. For typical edge networks, $d$ is very large, creating a communication bottleneck in bandwidth-constrained environments. Our goal is to retain NTK-DFL's convergence speed while reducing the number of communication rounds through training stabilization and, when required, reducing per-round Jacobian traffic through projection-based compression.

\subsection{Proposed SPARK Framework}
\label{subsec:proposed}

We present SPARK, a Stage-wise Projected NTK and Accelerated Regularization method that stabilizes momentum-accelerated NTK updates under heterogeneity through a co-design of temporal stabilization and convergence acceleration. The framework builds on two core components, stage-wise annealed distillation for training stabilization and momentum for convergence acceleration, and additionally uses random projection as a communication-reduction strategy. Figure~\ref{fig:spark_overview} illustrates the overall framework, and Algorithm~\ref{alg:spark} presents the complete procedure.

\textbf{Spatial compression via random projection.} Let $N_{\mathrm{lay}}$ denote the number of parameterized layers. The Jacobian is computed per layer via automatic differentiation. For each layer or parameter tensor $\ell$, client $i$ computes $\mathbf{J}_{i,\ell}^{(k)} \in \mathbb{R}^{N_i \times C \times o_\ell \times d_\ell^{\mathrm{in}}}$ using all assigned local samples, where $d_\ell^{\mathrm{in}}$ is the last parameter-axis dimension and $o_\ell$ denotes the product of the remaining parameter axes. Each layer-wise Jacobian is projected independently along this last axis. As a communication-reduction strategy, SPARK optionally compresses the Jacobian matrices via random projection while approximately preserving their spectral properties. For each layer $\ell$ with last-axis dimension $d_\ell^{\mathrm{in}}$, we set $k_\ell=\min(d_\ell^{\mathrm{in}}, k_{\mathrm{proj}})$, where $k_{\mathrm{proj}}$ denotes the projection cap. We then generate a deterministic random projection matrix $\mathbf{P}_\ell \in \mathbb{R}^{d_\ell^{\mathrm{in}} \times k_\ell}$ using a layer-specific seed derived from a global random seed. The projection matrix is sampled as:
\begin{equation}
\label{eq:projection_matrix}
\mathbf{P}_\ell \sim \mathcal{N}(0, k_\ell^{-1} \mathbf{I}), \quad \text{with entries} \quad P_{\ell,ij} \sim \mathcal{N}(0, 1/k_\ell),
\end{equation}
which ensures that the projection approximately preserves inner products with high probability, as guaranteed by the Johnson--Lindenstrauss lemma~\cite{johnson1984extensions}.

For an IoT device $i$, the compressed layer-wise Jacobian is computed as:
\begin{equation}
\label{eq:compression}
\tilde{\mathbf{J}}_{i,\ell}^{(k)} = \mathbf{J}_{i,\ell}^{(k)} \mathbf{P}_\ell \in \mathbb{R}^{N_i \times C \times o_\ell \times k_\ell}, \quad k_\ell=\min(d_\ell^{\mathrm{in}}, k_{\mathrm{proj}}).
\end{equation}

After flattening and concatenating the compressed layer-wise Jacobians along the parameter axes, the effective compressed dimension is $d_{\mathrm{proj}}=\sum_\ell o_\ell k_\ell$, while the uncompressed effective dimension is $d_{\mathrm{full}}=\sum_\ell o_\ell d_\ell^{\mathrm{in}}$. The number of transmitted Jacobian entries is reduced from $N_i C d_{\mathrm{full}}$ to $N_i C d_{\mathrm{proj}}$, achieving a compression ratio:
\begin{equation}
\label{eq:compression_ratio}
\rho =
\frac{
\sum_\ell o_\ell \min(d_\ell^{\mathrm{in}}, k_{\mathrm{proj}})
}{
\sum_\ell o_\ell d_\ell^{\mathrm{in}}
},
\end{equation}
where the communication saving depends on the layer shapes and the projection cap $k_{\mathrm{proj}}$. Crucially, the projection matrices $\mathbf{P}_\ell$ are generated deterministically from layer-specific seeds $s_\ell = \text{Hash}(s_{\text{global}} \| \text{name}_\ell)$, where $\|$ denotes concatenation and Hash is a cryptographic hash function. This ensures that all IoT devices use identical projection matrices for the same layer, enabling correct kernel computation, and that the projection is consistent across all communication rounds, preventing approximation drift. No additional communication is needed to synchronize the projection matrices. Setting $\mathbf{P}_\ell=\mathbf{I}$ recovers the full-Jacobian instantiation used in the main convergence, robustness, and ablation experiments; projection-related results are reported separately in Appendix~\ref{app:comm_cost}.

After receiving compressed Jacobians $\{\tilde{\mathbf{J}}_j^{(k)}\}_{j \in \mathcal{N}_i \cup \{i\}}$ from neighbors, IoT device $i$ constructs an aggregated Jacobian by concatenating along the sample dimension:
\begin{equation}
\label{eq:aggregate}
\tilde{\mathbf{J}}_{\text{agg}}^{(k)} = [\tilde{\mathbf{J}}_i^{(k)}; \tilde{\mathbf{J}}_{j_1}^{(k)}; \ldots; \tilde{\mathbf{J}}_{j_{|\mathcal{N}_i|}}^{(k)}] \in \mathbb{R}^{N_{\text{agg}} \times C \times d_{\mathrm{proj}}},
\end{equation}
where $N_{\text{agg}} = \sum_{j \in \mathcal{N}_i \cup \{i\}} N_j$ is the total number of samples from the IoT device and its neighbors. The empirical Neural Tangent Kernel is then computed as:
\begin{equation}
\label{eq:kernel}
\mathbf{K}^{(k)} = \tilde{\mathbf{J}}_{\text{agg}}^{(k)} (\tilde{\mathbf{J}}_{\text{agg}}^{(k)})^\top \in \mathbb{R}^{(N_{\text{agg}} \cdot C) \times (N_{\text{agg}} \cdot C)}.
\end{equation}

This kernel matrix encodes the geometry of the loss landscape in the compressed parameter space, where each entry $K_{ij}^{(k)} = \langle \tilde{\mathbf{J}}_{\text{agg},i}^{(k)}, \tilde{\mathbf{J}}_{\text{agg},j}^{(k)} \rangle$ measures the similarity between training sample pairs. Although $\tilde{\mathbf{J}}_{\text{agg}}^{(k)}$ approximates the full Jacobian, the kernel $\mathbf{K}^{(k)}$ provably retains the spectral structure needed for convergence.

\textbf{Temporal stabilization via stage-wise annealed distillation.} To stabilize momentum-accelerated NTK updates, and to counteract sketching noise when projection is enabled, SPARK uses a stage-wise soft-label regularizer. After neighborhood weight averaging, each client evaluates the aggregated neighborhood data using its local consensus model. The resulting soft labels provide smoothed neighborhood-consensus targets that compensate for noisy updates while maintaining training stability.

The distillation operates in two phases. During the initial warm-up period spanning rounds $k \in \{1, \ldots, R_{\text{warm}}\}$, we use only ground-truth hard labels as training targets with mixing coefficient $\alpha^{(k)} = 1.0$ and temperature parameter $\tau^{(k)} = 1.0$. This lets the NTK update form a good initialization without interference from noisy soft targets. After the warm-up phase, we gradually transition to a mixture of hard and soft labels. The mixing coefficient $\alpha^{(k)}$ follows a cosine annealing schedule, while the temperature parameter $\tau^{(k)}$ increases linearly to progressively soften the probability distributions.

Specifically, for a given round $k > R_{\text{warm}}$, we compute the progress ratio:
\begin{equation}
\label{eq:progress}
p^{(k)} = \frac{k - R_{\text{warm}}}{R - R_{\text{warm}}} \in [0,1],
\end{equation}
where $R_{\text{warm}}$ is the number of warm-up rounds and $R$ is the total number of rounds. The mixing coefficient and temperature are then updated as:
\begin{equation}
\label{eq:alpha_schedule}
\alpha^{(k)} = \alpha_{\text{final}} + \frac{1}{2}(\alpha_{\text{init}} - \alpha_{\text{final}})(1 + \cos(\pi p^{(k)})),
\end{equation}
\begin{equation}
\label{eq:tau_schedule}
\tau^{(k)} = \tau_{\text{init}} + (\tau_{\text{final}} - \tau_{\text{init}}) p^{(k)}.
\end{equation}

The soft labels are constructed by evaluating the local consensus model on the data of each neighbor $j \in \mathcal{N}_i \cup \{i\}$ to obtain logits $\mathbf{z}_j^{(k)} \in \mathbb{R}^{N_j \times C}$, which are concatenated to form $\mathbf{z}_{\text{agg}}^{(k)} \in \mathbb{R}^{N_{\text{agg}} \times C}$, then applying temperature-scaled softmax:
\begin{equation}
\label{eq:soft_labels}
Y_{\text{soft},nc}^{(k)} = \frac{\exp(z_{\text{agg},nc}^{(k)} / \tau^{(k)})}{\sum_{c'=1}^{C} \exp(z_{\text{agg},nc'}^{(k)} / \tau^{(k)})}.
\end{equation}
The final target for kernel evolution is:
\begin{equation}
\label{eq:distillation_target}
\mathbf{Y}_{\text{tgt}}^{(k)} = \alpha^{(k)} \mathbf{Y}_{\text{hard}} + (1 - \alpha^{(k)}) \mathbf{Y}_{\text{soft}}^{(k)},
\end{equation}
where $\mathbf{Y}_{\text{hard}} \in \{0,1\}^{N_{\text{agg}} \times C}$ denotes the one-hot encoded ground-truth labels. This annealing reduces variance by smoothing the noisy update directions induced by momentum and, when enabled, projection, while enabling implicit knowledge sharing across the graph.

\textbf{Convergence acceleration via momentum optimization.} Beyond compression and stabilization, SPARK incorporates momentum-based optimization to accelerate convergence. Traditional NTK-based methods evolve predictions through direct kernel-based updates, which can converge slowly under high data heterogeneity. SPARK addresses this by maintaining a velocity term that accumulates gradient information across rounds.

Specifically, let $\mathbf{w}_i^{(k)}$ denote IoT device $i$'s model weights at round $k$, and let $\Delta \mathbf{w}_{i}^{(k)}$ be the weight update computed from kernel-based prediction evolution (detailed below). We maintain a velocity vector $\mathbf{v}_i^{(k)} \in \mathbb{R}^d$ that is updated as:
\begin{equation}
\label{eq:velocity_update}
\mathbf{v}_i^{(k+1)} = \mu \mathbf{v}_i^{(k)} + \Delta \mathbf{w}_{i}^{(k)},
\end{equation}
where $\mu \in [0,1)$ is the momentum coefficient. The momentum term $\mu \mathbf{v}_i^{(k)}$ accumulates past gradient information, letting the optimization keep its direction through noisy updates and accelerate along consistent descent directions.

SPARK employs Nesterov accelerated momentum~\cite{nesterov2004introductory}, which has stronger convergence guarantees than standard momentum. The Nesterov update applies a lookahead step:
\begin{equation}
\label{eq:nesterov_update}
\mathbf{w}_i^{(k+1)} = \mathbf{w}_i^{(k)} + \mu \mathbf{v}_i^{(k+1)} + \Delta \mathbf{w}_{i}^{(k)}.
\end{equation}

We set $\mu = 0.9$, which accelerates convergence while preserving stability.

\textbf{Kernel-based prediction evolution and weight updates.} Given the empirical kernel $\mathbf{K}^{(k)}$ and target labels $\mathbf{Y}_{\text{tgt}}^{(k)}$, we evolve the network predictions via kernel-based gradient descent. Starting from the current model's predictions $\mathbf{f}_0^{(k)} = f(\mathbf{X}_{\text{agg}}; \mathbf{w}_i^{(k)})$, the prediction trajectory is computed as:
\begin{equation}
\label{eq:prediction_evolution}
\mathbf{f}_t^{(k)} = \mathbf{f}_0^{(k)} - \eta \sum_{s=0}^{t-1} \mathbf{K}^{(k)} \nabla_{\mathbf{f}} \mathcal{L}(\mathbf{f}_s^{(k)}, \mathbf{Y}_{\text{tgt}}^{(k)}),
\end{equation}
where $\eta$ is the learning rate and $\nabla_{\mathbf{f}} \mathcal{L}$ is the gradient with respect to predictions. For cross-entropy loss, this reduces to:
\begin{equation}
\label{eq:prediction_evolution_ce}
\mathbf{f}_t^{(k)} = \mathbf{f}_0^{(k)} + \eta \mathbf{K}^{(k)} \left( t\mathbf{Y}_{\text{tgt}}^{(k)} - \sum_{s=0}^{t-1} \text{softmax}(\mathbf{f}_s^{(k)}) \right).
\end{equation}

Following NTK-DFL, we evaluate a user-specified grid of
evolution timesteps $\mathcal{T}$ and select the candidate with the lowest
loss on the aggregated neighborhood data,
$t^\star=\arg\min_{t\in\mathcal{T}} \mathcal{L}_{\mathrm{agg}}(\mathbf{f}_t^{(k)},\mathbf{Y}_{\mathrm{agg}})$,
where $\mathcal{T}=\{100,200,\ldots,800\}$ and $\mathbf{Y}_{\mathrm{agg}}$
denotes the aggregated ground-truth labels of the neighborhood. After obtaining the optimal prediction trajectory, we compute weight updates in the compressed space:
\begin{equation}
\label{eq:compressed_update}
\Delta \tilde{\mathbf{w}}_{t^*} = -\frac{\eta}{N_{\text{agg}}} (\tilde{\mathbf{J}}_{\text{agg}}^{(k)})^\top \left( \sum_{s=0}^{t^*-1} \text{softmax}(\mathbf{f}_s^{(k)}) - t^* \mathbf{Y}_{\text{tgt}}^{(k)} \right),
\end{equation}
where $\Delta \tilde{\mathbf{w}}_{t^*} \in \mathbb{R}^{d_{\mathrm{proj}}}$ is the gradient in the compressed subspace. To obtain the update in the original parameter space, we apply the back-projection:
\begin{equation}
\label{eq:backproject}
\Delta \mathbf{w}_{t^*} = \mathbf{P} \Delta \tilde{\mathbf{w}}_{t^*} \in \mathbb{R}^{d_{\mathrm{full}}}.
\end{equation}

This back-projection does not invert the random projection when
$d_{\mathrm{proj}}<d_{\mathrm{full}}$; the resulting update equals the
sketched direction $\Delta \mathbf w_{t^*}=-\eta\,PP^\top g_{t^*}^{(k)}$,
where $g_{t^*}^{(k)}$ denotes the full NTK gradient direction. Its
unbiasedness over the projection randomness and its descent property are
formalized in \cref{def:proj_ntk} and Lemma~\ref{lem:backproj_descent},
with the residual $(I_{d_{\mathrm{full}}}-PP^\top)g_{t^*}^{(k)}$ absorbed
into the bias term $\delta_{\mathrm{total}}$.

The final model update incorporates momentum as described in \cref{eq:velocity_update,eq:nesterov_update}.

\section{Theoretical Analysis}
\label{sec:theory}
We provide a lightweight non-convex convergence guarantee for SPARK. The bound decomposes into an approximation bias term induced by sketched-NTK back-projection and a variance term controlled by the stage-wise annealed distillation schedule. 

\noindent\textbf{Global objective.} Let $\mathcal L(w)=\frac{1}{M}\sum_{i=1}^M \mathcal L_i(w)$ be the global objective. At round $k$, each IoT device computes a back-projected update $\Delta w_i^{(k)}$ and performs the (Nesterov-style) momentum step
\begin{equation}
\label{eq:spark_update_main}
\begin{aligned}
v_i^{(k+1)} &= \mu v_i^{(k)} + \Delta w_i^{(k)},\\
w_i^{(k+1)} &= w_i^{(k)} + \mu v_i^{(k+1)} + \Delta w_i^{(k)}.
\end{aligned}
\end{equation}

where $\mu\in[0,1)$. Define averages $\bar w^{(k)}=\frac{1}{M}\sum_{i=1}^M w_i^{(k)}$, $\bar v^{(k)}=\frac{1}{M}\sum_{i=1}^M v_i^{(k)}$, and $\bar\Delta^{(k)}=\frac{1}{M}\sum_{i=1}^M \Delta w_i^{(k)}$.

\begin{definition}
\label{def:proj_ntk}

(Sketched NTK direction). Let
\[
P=\mathrm{blkdiag}\!\left(
I_{o_1}\otimes P_1,\ldots,
I_{o_{N_{\mathrm{lay}}}}\otimes P_{N_{\mathrm{lay}}}
\right)
\in\mathbb R^{d_{\mathrm{full}}\times d_{\mathrm{proj}}},
\]
where
$d_{\mathrm{proj}}=\sum_{\ell=1}^{N_{\mathrm{lay}}}o_\ell k_\ell$,
$d_{\mathrm{full}}=\sum_{\ell=1}^{N_{\mathrm{lay}}}o_\ell d_\ell^{\mathrm{in}}$, and
$k_\ell=\min(d_\ell^{\mathrm{in}},k_{\mathrm{proj}})$. The matrix $P$ is
fixed and shared by all IoT devices. Define the sketching operator
\begin{equation}
\label{eq:sketch_op}
S_P \triangleq PP^\top
\in\mathbb R^{d_{\mathrm{full}}\times d_{\mathrm{full}}}.
\end{equation}
SPARK's back-projected update at each device uses the sketched gradient
direction $S_P\,g_i(w)$ in place of the full NTK direction $g_i(w)$.
With the layer-wise Gaussian normalization in \cref{eq:projection_matrix},
$\mathbb E_P[S_P]=I_{d_{\mathrm{full}}}$, so $S_P$ is an unbiased sketch of the identity
before fixing the projection seed. After the seed is fixed, the
realized approximation error is absorbed into the bias term
$\delta_{\mathrm{total}}$ (Assumption~\ref{ass:bias}).

\end{definition}

\begin{definition}
\label{def:distill}
(Annealed distillation objective). Recalling the distillation target
$Y^{(k)}_{\mathrm{tgt}}$ in \cref{eq:distillation_target} with schedule
$\alpha^{(k)}\in[0,1]$, $\tau^{(k)}\ge 1$, and writing
$p_{\tau^{(k)}}(w)\triangleq \mathrm{softmax}(f(x;w)/\tau^{(k)})$,
define the effective local objective
\begin{equation}
\label{eq:distill_obj}
\begin{aligned}
\mathcal L_i^{(k)}(w)
\triangleq\;&
\alpha^{(k)}\,\mathcal L_i^{\mathrm{CE}}(w)
+\bigl(1-\alpha^{(k)}\bigr)\,\bigl(\tau^{(k)}\bigr)^2 \\
&\cdot \mathrm{KL}\!\Bigl(
Y^{(k)}_{\mathrm{soft}}
\,\big\|\, p_{\tau^{(k)}}(w)
\Bigr).
\end{aligned}
\end{equation}
The update $\Delta w_i^{(k)}$ is computed via sketched-NTK evolution using
$\bigl(Y^{(k)}_{\mathrm{tgt}},P\bigr)$ and then back-projected to $\mathbb R^{d_{\mathrm{full}}}$.
\end{definition}

\begin{assumption}
\label{ass:regular}
(Smoothness and annealed stochasticity). For all IoT devices $i$ and rounds $k$, we assume $\mathcal L$ and each $\mathcal L_i^{(k)}$ are $L$-smooth. Moreover, there exist $\sigma^2>0$ and a non-increasing sequence $\{\gamma_k\}_{k\ge 0}\subset(0,1]$ such that, with filtration $\mathcal F_k$,
\begin{equation}
\label{eq:annealed_var}
\mathbb E\!\left[\left\|\Delta w_i^{(k)}-\mathbb E[\Delta w_i^{(k)}\mid\mathcal F_k]\right\|^2\Bigm|\mathcal F_k\right]
\le \eta^2\sigma^2\gamma_k .
\end{equation}
Conditioned on $\mathcal F_k$, the update noises are uncorrelated across IoT devices.
\end{assumption}

\begin{assumption}
\label{ass:bias}

(Sketched-NTK approximation bias).
Let $\bar\Delta_{\mathbb E}^{(k)}\triangleq \mathbb E[\bar\Delta^{(k)}\mid\mathcal F_k]$. Define the momentum virtual sequence
\begin{equation}
\label{eq:virtual_seq}
\bar z^{(k)}\triangleq \bar w^{(k)}+\frac{\mu^2}{1-\mu}\bar v^{(k)}.
\end{equation}
We assume there exists $\delta_{\mathrm{total}}\ge 0$ such that for all $k$,
\begin{equation}
\label{eq:bias_bound}
\left\|\bar\Delta_{\mathbb E}^{(k)}+\eta\nabla\mathcal L(\bar z^{(k)})\right\|
\le \eta\,\delta_{\mathrm{total}}.
\end{equation}
The aggregate bias $\delta_{\mathrm{total}}$ admits the decomposition
\begin{equation}
\label{eq:delta_decomp}
\delta_{\mathrm{total}}
\;\le\;
\delta_{\mathrm{proj}}(k_{\mathrm{proj}})
\;+\;
\delta_{\mathrm{ntk}}
\;+\;
\delta_{\mathrm{graph}},
\end{equation}
where $\delta_{\mathrm{proj}}(k_{\mathrm{proj}})$ arises from replacing the full NTK
gradient $g$ with the sketched direction $PP^\top g$ (\cref{def:proj_ntk}),
$\delta_{\mathrm{ntk}}$ accounts for finite-width NTK approximation error, and
$\delta_{\mathrm{graph}}$ captures decentralized neighbor-aggregation mismatch.

\end{assumption}

\begin{theorem}
\label{thm:spark}
(Convergence of SPARK (non-convex)). Under Assumptions~\ref{ass:regular}--\ref{ass:bias}, assume the stepsize satisfies
\begin{equation}
\label{eq:stepsize_cond}
\eta \le \frac{1-\mu}{4L}.
\end{equation}
Let $\bar\gamma \triangleq \frac{1}{T}\sum_{k=0}^{T-1}\gamma_k$. Then after $T$ rounds,
\begin{equation}
\label{eq:main_bound}
\begin{aligned}
\min_{0\le k<T}\mathbb E\|\nabla \mathcal L(\bar z^{(k)})\|^2
\le\;&
\frac{2(1-\mu)\big(\mathcal L(\bar z^{(0)})-\mathcal L^\star\big)}{\eta T} \\
&+
\frac{C_1 L\eta\sigma^2}{M(1-\mu)}\,\bar\gamma
+
C_2\,\delta_{\mathrm{total}}^2,
\end{aligned}
\end{equation}
where the two constants are explicit and absolute,
\begin{equation}
\label{eq:constants_main}
C_1=1,
\qquad
C_2=2+\frac{2L\eta}{1-\mu}\;\le\;\frac{5}{2},
\end{equation}
the upper bound on $C_2$ following from the stepsize condition
\cref{eq:stepsize_cond} (which gives $\tfrac{2L\eta}{1-\mu}\le\tfrac12$).
Both are independent of $T$, $M$ and $\sigma^2$; their derivation by
coefficient matching is given in \cref{eq:constants_explicit} of
Appendix~\ref{app:math_analysis}. Here
$\mathcal L^\star$ denotes a finite lower bound of the
global objective.
\end{theorem}

\begin{corollary}
\label{cor:tradeoff}
(Communication-accuracy trade-off). Choosing
$\eta=\Theta(1/\sqrt{T})$ subject to
\cref{eq:stepsize_cond} yields
\begin{equation}
\label{eq:cor_tradeoff_rate}
\min_{0\le k<T}\mathbb E\bigl\|\nabla \mathcal L(\bar z^{(k)})\bigr\|^2
=
\mathcal O\!\left(
\frac{1}{\sqrt{T}}
+
\frac{\bar\gamma}{M\sqrt{T}}
+
\delta_{\mathrm{total}}^2
\right).
\end{equation}
For the projection-enabled instantiation, SPARK communicates compressed Jacobian objects with effective dimension $d_{\mathrm{proj}}$ instead of $d_{\mathrm{full}}$; hence the compression ratio follows \cref{eq:compression_ratio}.
\end{corollary}

From \cref{eq:cor_tradeoff_rate}, the convergence bound comprises optimization and stochastic terms that decrease with $T$, together with an irreducible approximation floor $\delta_{\mathrm{total}}^2$. Increasing the number of communication rounds therefore cannot eliminate the errors represented by $\delta_{\mathrm{total}}$.

\begin{remark}
More specifically, the bias floor $\delta_{\mathrm{total}}^2$ aggregates three sources, namely
random projection distortion, finite-width NTK approximation error, and
graph-induced mismatch from decentralized aggregation. The projection
component is controlled by the projection cap $k_{\mathrm{proj}}$, which
determines the layer-wise projected dimensions
$\{k_\ell\}_{\ell=1}^{N_{\mathrm{lay}}}$
and the total effective dimension $d_{\mathrm{proj}}$, under
Johnson--Lindenstrauss concentration. The graph component depends on
the communication topology. In this work, we treat $\delta_{\mathrm{total}}$ 
as a model-agnostic constant in the convergence analysis. Characterizing 
its explicit dependence on $k_{\mathrm{proj}}$ and graph topology is left to future work.
\end{remark}


\section{Experiments}
\label{sec:experiments}

\subsection{Experimental Setup}

\textbf{Datasets and Model}\quad Following Thompson et al.~\cite{ntkdfl}, we evaluate SPARK on three datasets: Fashion-MNIST~\cite{fashionmnist}, 
FEMNIST~\cite{leaf}, and MNIST~\cite{mnist}. Fashion-MNIST 
contains 60,000 training and 10,000 test samples across 10 categories, 
serving as our primary benchmark. FEMNIST provides naturally heterogeneous 
data based on handwriting styles, while MNIST allows controlled
heterogeneity. Data heterogeneity is introduced via symmetric Dirichlet 
distribution with concentration parameter $\alpha$~\cite{dirichlet}, where 
a smaller $\alpha$ yields stronger label skew. For each IoT device $i$, we 
sample $\mathbf{q}_i \sim \text{Dir}(\alpha \mathbbm{1})$ with 
$\mathbf{q}_i \in \mathbb{R}^{C}$ defining label probabilities for $C$ 
classes. Following NTK-DFL, the Dirichlet client partition is
generated once before training with a fixed partition seed and kept fixed
across all communication rounds, with no per-round data resampling. The model architecture is a two-layer MLP with 100 hidden neurons, 
yielding $d = 79{,}510$ parameters.

\begin{figure*}[t]
    \centering
    \begin{subfigure}[b]{0.35\textwidth}
        \centering
        \includegraphics[width=\linewidth]{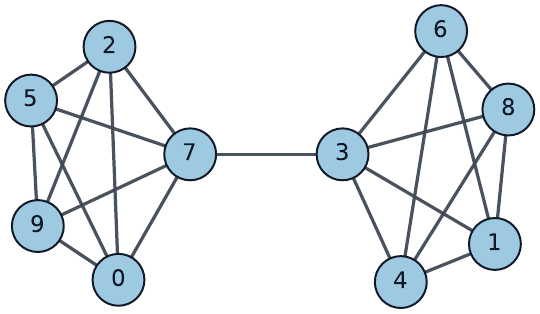}
        \caption{Cluster}
    \end{subfigure}
    \hfill
    \begin{subfigure}[b]{0.30\textwidth}
        \centering
        \includegraphics[width=\linewidth]{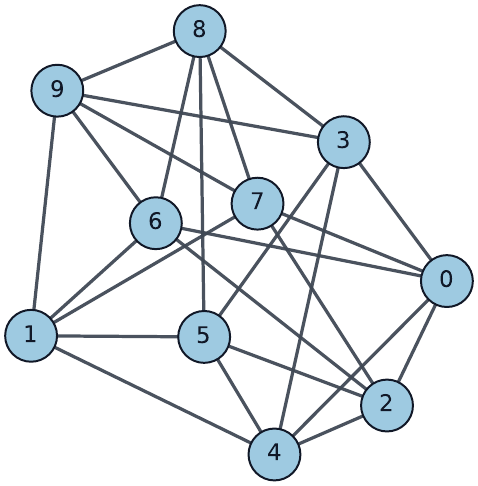}
        \caption{Regular ($\kappa=5$)}
    \end{subfigure}
    \hfill
    \begin{subfigure}[b]{0.30\textwidth}
        \centering
        \includegraphics[width=\linewidth]{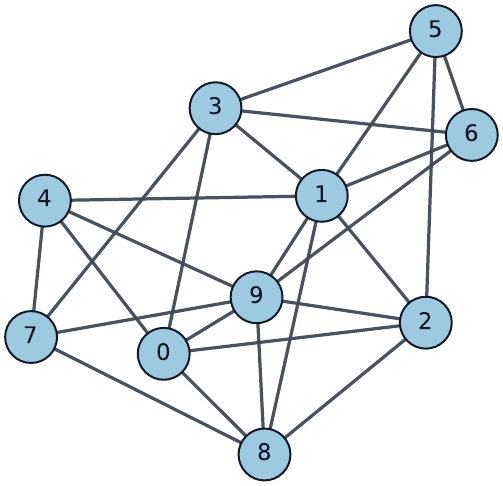}
        \caption{Erd\H{o}s--R\'enyi ($\kappa=5$)}
    \end{subfigure}

    \vspace{2mm}

    \begin{subfigure}[b]{0.30\textwidth}
        \centering
        \includegraphics[width=\linewidth]{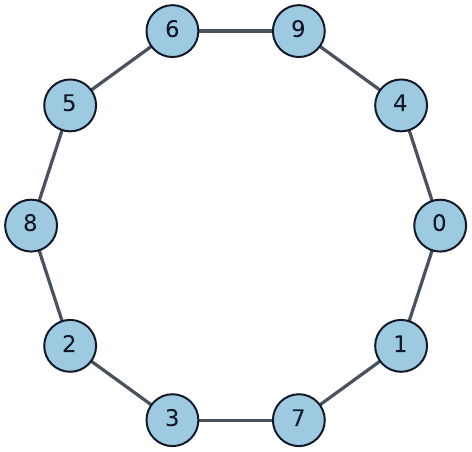}
        \caption{Ring}
    \end{subfigure}
    \hspace{0.04\textwidth}
    \begin{subfigure}[b]{0.50\textwidth}
        \centering
        \includegraphics[width=\linewidth]{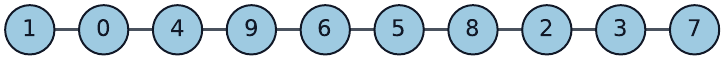}
        \caption{Line}
    \end{subfigure}
    \caption{The five communication topologies used in the robustness evaluation.}
    \label{fig:topology_gallery}
\end{figure*}

\textbf{Network Topology}\quad We employ a sparse, time-variant 
$\kappa$-regular graph with $\kappa = 5$ across $M = 300$ IoT devices. At 
each communication round, a new random graph with the same degree is 
generated. This sparse connectivity reflects realistic
decentralized settings with limited peer connectivity and ensures a fair
comparison across methods. We also evaluate various topologies including cluster, 
regular, Erd\H{o}s--R\'enyi, ring, and line graphs to assess robustness to 
network structure, as illustrated
in Fig.~\ref{fig:topology_gallery}. These range from densely clustered and
uniform-degree graphs to randomly connected and sparsely chained
structures with degree two.

\textbf{Baseline Methods.} We compare SPARK against representative 
decentralized federated learning methods, including D-PSGD~\cite{d-psgd}, 
DFedAvg, DFedAvgM~\cite{dfedavg}, DFedSAM~\cite{dfedsam}, 
and NTK-DFL~\cite{ntkdfl}. These methods cover both first-order gradient 
and higher-order NTK-based approaches.

\textbf{Evaluation Metrics.} We measure test accuracy on a global 
holdout set to evaluate model generalization. Following Thompson et 
al.~\cite{ntkdfl}, we assess both aggregated model accuracy, computed 
by averaging client models, and average individual client accuracy. 
This dual evaluation exposes the generalization gap between
the global consensus and local models, which is informative in heterogeneous
settings.
The convergence, robustness, and ablation
experiments use the full-Jacobian instantiation of SPARK by setting
$\mathbf{P}_\ell=\mathbf{I}$. Projection-related communication results are
reported separately in Appendix~\ref{app:comm_cost}.
Detailed hyperparameter settings are provided in Appendix~\ref{app:hyperparams}.
Experiments are repeated over three independent random trials on a single NVIDIA RTX 5090 GPU, and results are reported as mean and standard deviation. 

\subsection{Convergence Speed and Final Accuracy}

\begin{figure*}[!t]
    \centering
    \includegraphics[width=0.48\linewidth]{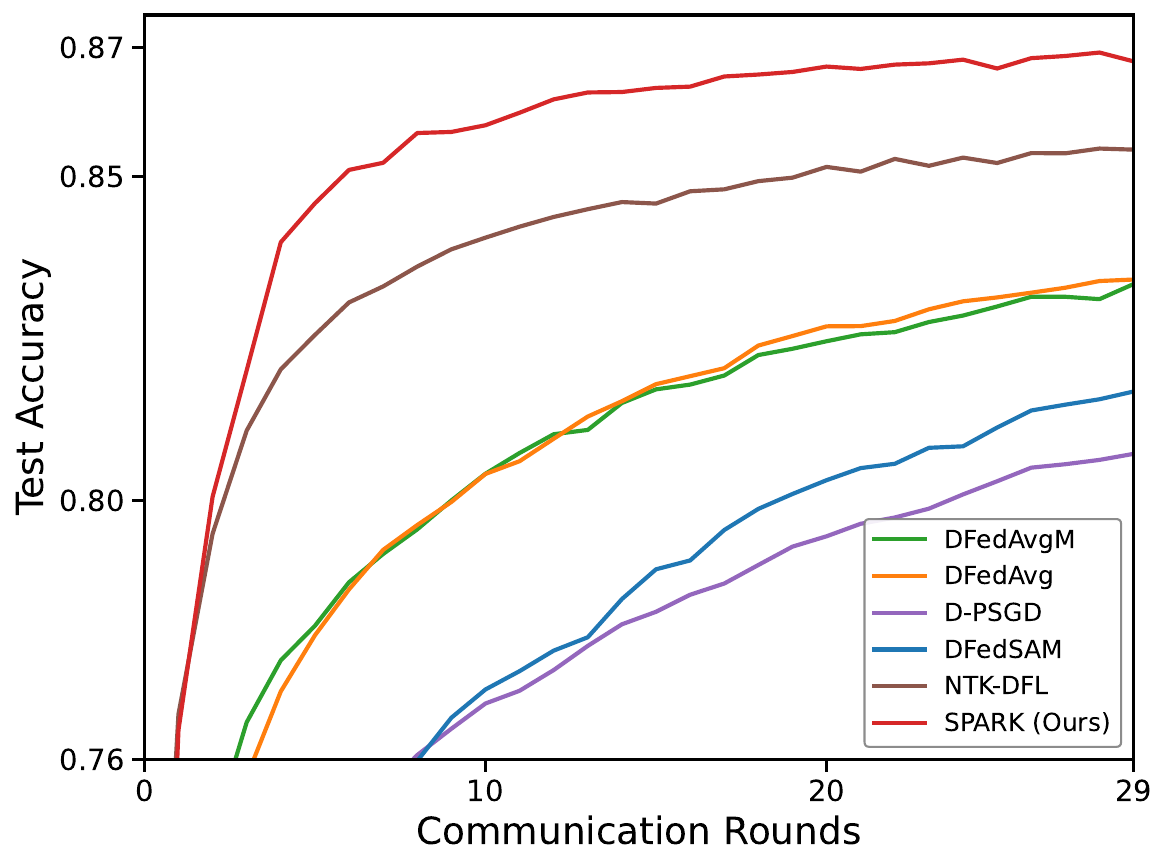}
    \hfill 
    \includegraphics[width=0.48\linewidth]{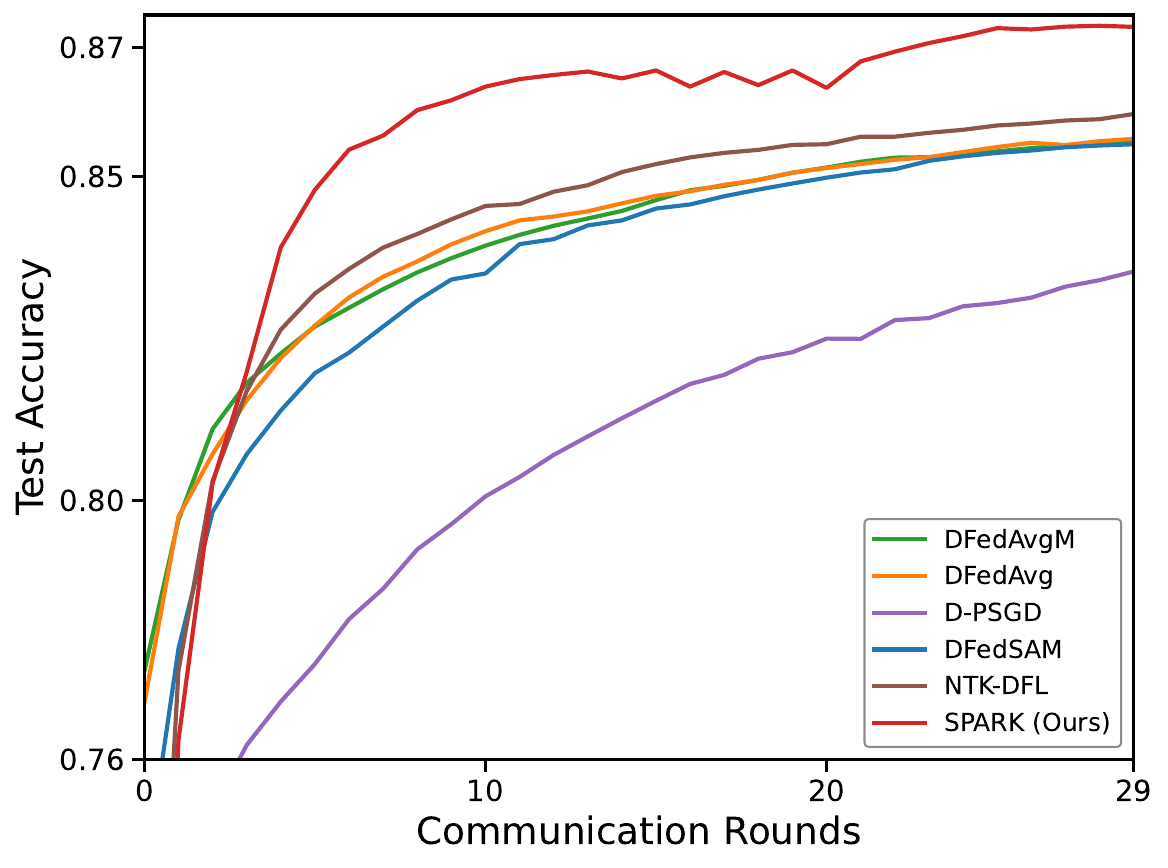}
    
    \caption{Convergence of different methods on Fashion-MNIST under 
    non-IID ($\alpha=0.1$) and IID settings. Curves show mean performance 
    over three independent runs.}
    \vspace{-2mm}
    \label{fig:topology_compare}
\end{figure*}

\begin{figure*}[!t]
    \centering
    \begin{subfigure}[b]{0.32\textwidth}
        \centering
        \includegraphics[width=\linewidth]{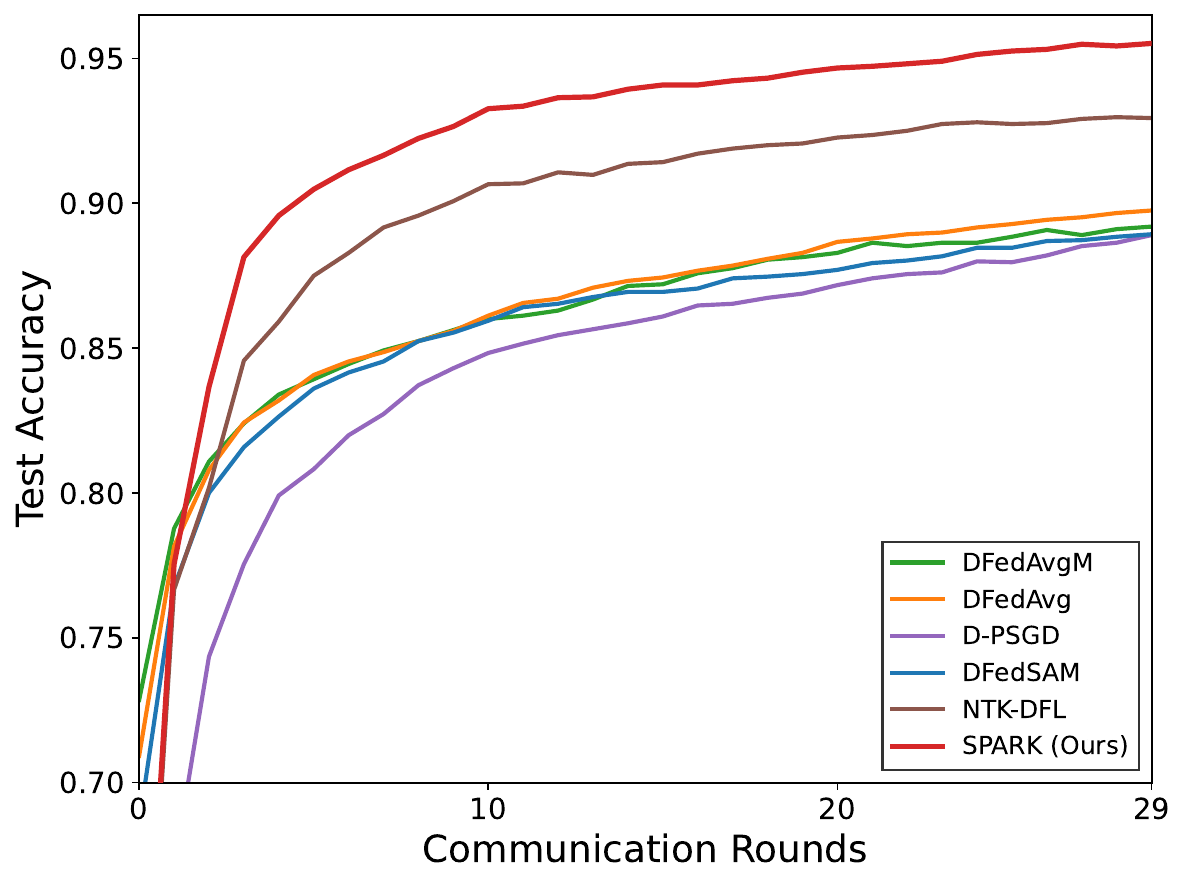}
        \caption{FEMNIST (Feature-skewed)}
    \end{subfigure}
    \hfill 
    \begin{subfigure}[b]{0.32\textwidth}
        \centering
        \includegraphics[width=\linewidth]{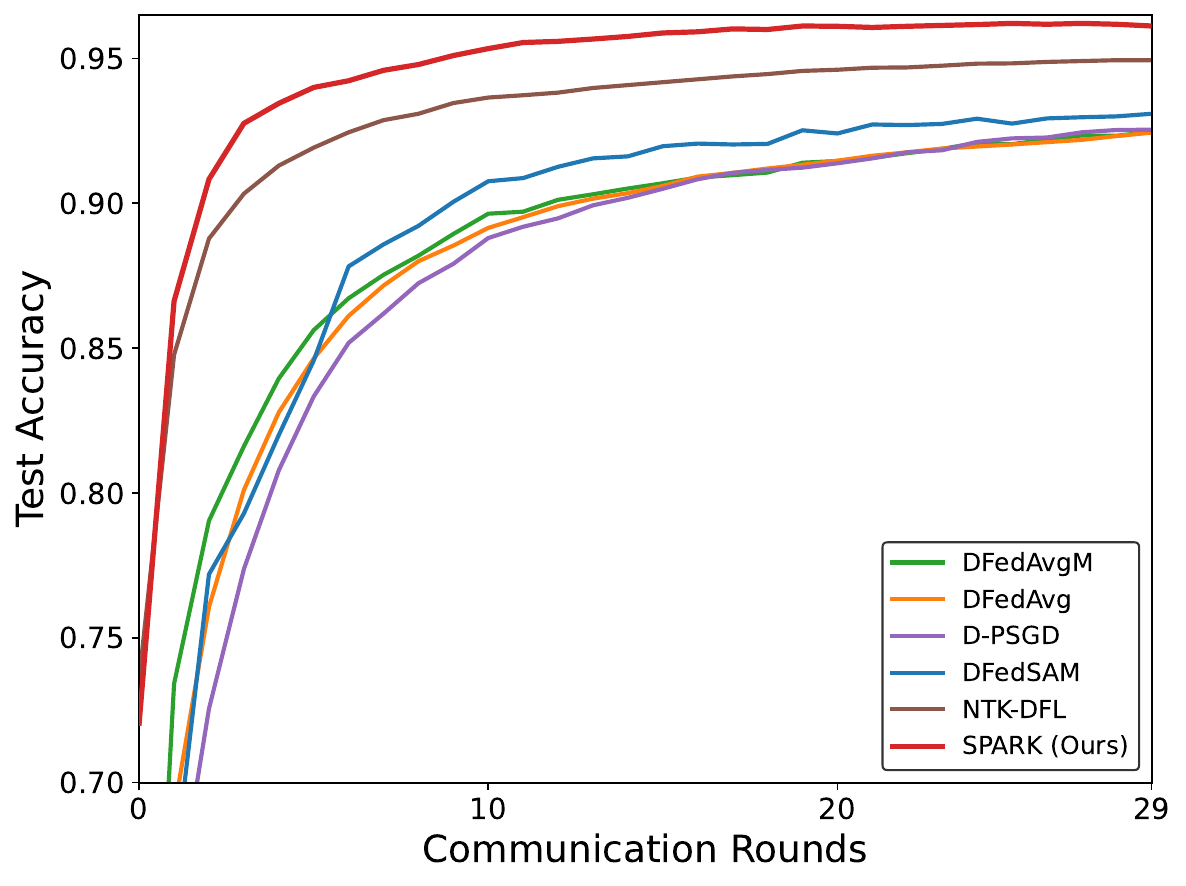}
        \caption{MNIST $(\alpha = 0.05)$}
    \end{subfigure}
    \hfill 
    \begin{subfigure}[b]{0.32\textwidth}
        \centering
        \includegraphics[width=\linewidth]{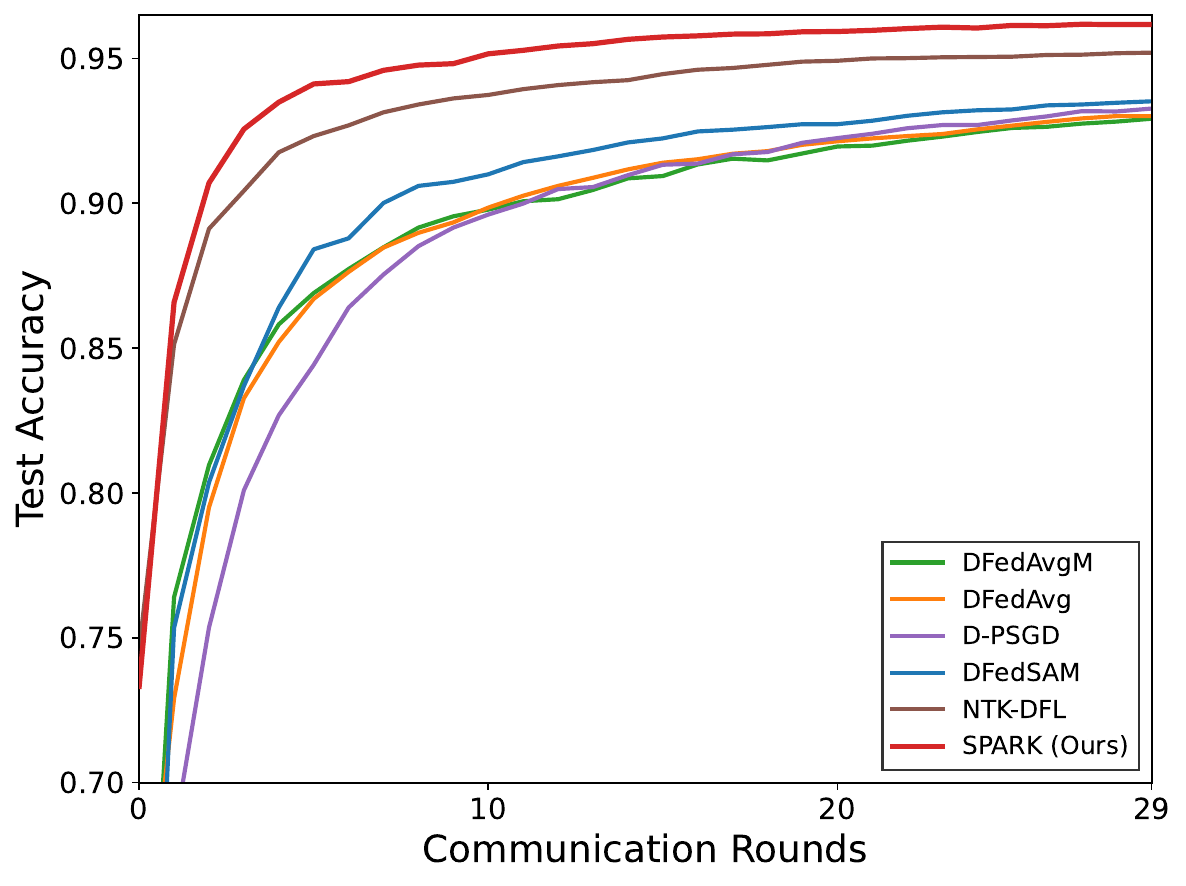}
        \caption{MNIST $(\alpha = 0.1)$}
    \end{subfigure}

    \caption{Convergence of various methods on heterogeneous datasets: 
    (a) FEMNIST (feature-skewed), (b) MNIST ($\alpha=0.05$), and 
    (c) MNIST ($\alpha=0.1$).}
    \label{fig:different_dataset_convergence}
\end{figure*}

SPARK converges faster than all baselines across settings. Figure~\ref{fig:topology_compare} illustrates convergence trajectories on Fashion-MNIST under highly heterogeneous ($\alpha=0.1$) and IID settings. SPARK's advantage grows with heterogeneity. Under high heterogeneity with $\alpha=0.1$, SPARK achieves an early accuracy gain of about 2.0\% to 2.3\% over NTK-DFL within the first few communication rounds, and remains consistently better throughout training. Table~\ref{tab:convergence_rounds} quantifies communication rounds required to reach 85\% test accuracy. SPARK reduces the rounds to 85\% accuracy to 6 across all heterogeneity levels, yielding up to 3 times faster convergence than NTK-DFL and up to 12 times fewer rounds than DFedAvg. In this full-Jacobian convergence comparison, the reduction in rounds translates to about 70\% lower total communication than NTK-DFL to reach 85\% accuracy under $\alpha=0.1$. Figure~\ref{fig:different_dataset_convergence} shows that SPARK exhibits comparable convergence advantages on feature-skewed FEMNIST and label-skewed MNIST, while achieving final test accuracy improvements ranging from 1\% to 2.6\% over the best-performing baseline.

We conduct two-sided paired $t$-tests of SPARK against NTK-DFL on the
round-29 aggregated accuracy over three shared seeds, and report the
paired effect size $d_z=\bar{\Delta}/s_{\Delta}$, which is not comparable
to between-group Cohen's $d$. As shown in Table~\ref{tab:significance},
all $95\%$ confidence intervals exclude zero and all $p<0.01$, so the
gains are significant at the $95\%$ level across all heterogeneity levels.

\begin{table*}[!t]
    \scriptsize
    \centering
    \begin{minipage}[t]{0.48\textwidth}
        \vspace{0pt}
        \centering
        \caption{Communication rounds to reach 85\% test accuracy on 
Fashion-MNIST under varying heterogeneity levels.}
        \label{tab:convergence_rounds}
        \vspace{2mm}
        \begin{tabular*}{\linewidth}{@{\extracolsep{\fill}}lccc@{}}
                \toprule
                \multirow{2}{*}{\textbf{Method}} & \multicolumn{3}{c}{\textbf{Heterogeneity Level} ($\alpha$)} \\
                \cmidrule(lr){2-4}
                 & \textbf{IID} & $\alpha = 0.5$ & $\alpha = 0.1$ \\
                \midrule
                DFedAvg  & 19 & 36 & 73 \\
                DFedAvgM & 19 & 38 & 79 \\
                DFedSAM  & 22 & 42 & 200+ \\
                D-PSGD   & 56 & 82 & 148 \\
                NTK-DFL  & 14 & 16 & 19 \\
                \textbf{SPARK} & \textbf{6} & \textbf{6} & \textbf{6} \\
                \bottomrule
        \end{tabular*}
    \end{minipage}
    \hfill
    \begin{minipage}[t]{0.48\textwidth}
        \vspace{0pt}
        \centering
        \caption{Test accuracy (\%) on Fashion-MNIST under varying 
heterogeneity levels ($\alpha$). Smaller $\alpha$ indicates higher 
heterogeneity.}
        \label{tab:heterogeneity_robustness}
        \vspace{2mm}
        \setlength{\tabcolsep}{1.5pt}
        \begin{tabular*}{\linewidth}{@{\extracolsep{\fill}}lccc@{}}
                \toprule
                \multirow{2}{*}{\textbf{Method}} & \multicolumn{3}{c}{\textbf{Heterogeneity Level} ($\alpha$)} \\
                \cmidrule(lr){2-4}
                 & $\alpha = 0.1$  & $\alpha = 0.3$  & $\alpha = 0.5$  \\
                \midrule
                DFedAvg  & $83.41 \pm 0.29$ & $84.76 \pm 0.11$ & $84.75 \pm 0.13$ \\
                DFedAvgM & $83.34 \pm 0.15$ & $84.54 \pm 0.14$ & $84.61 \pm 0.21$ \\
                DFedSAM  & $81.68 \pm 0.39$ & $83.91 \pm 0.06$ & $84.42 \pm 0.06$ \\
                D-PSGD   & $80.72 \pm 0.21$ & $82.43 \pm 0.13$ & $82.47 \pm 0.19$ \\
                NTK-DFL  & $85.42 \pm 0.03$ & $85.57 \pm 0.02$ & $85.65 \pm 0.19$ \\
                \textbf{SPARK} & $\mathbf{86.78 \pm 0.02}$ & $\mathbf{86.63 \pm 0.05}$ & $\mathbf{86.42 \pm 0.14}$ \\
                \bottomrule
        \end{tabular*}
    \end{minipage}
\end{table*}

\subsection{Factor Analyses for SPARK}
\begin{figure}[!t]
    \centering
    \begin{subfigure}[b]{0.48\linewidth}
        \centering
        \includegraphics[width=\linewidth]{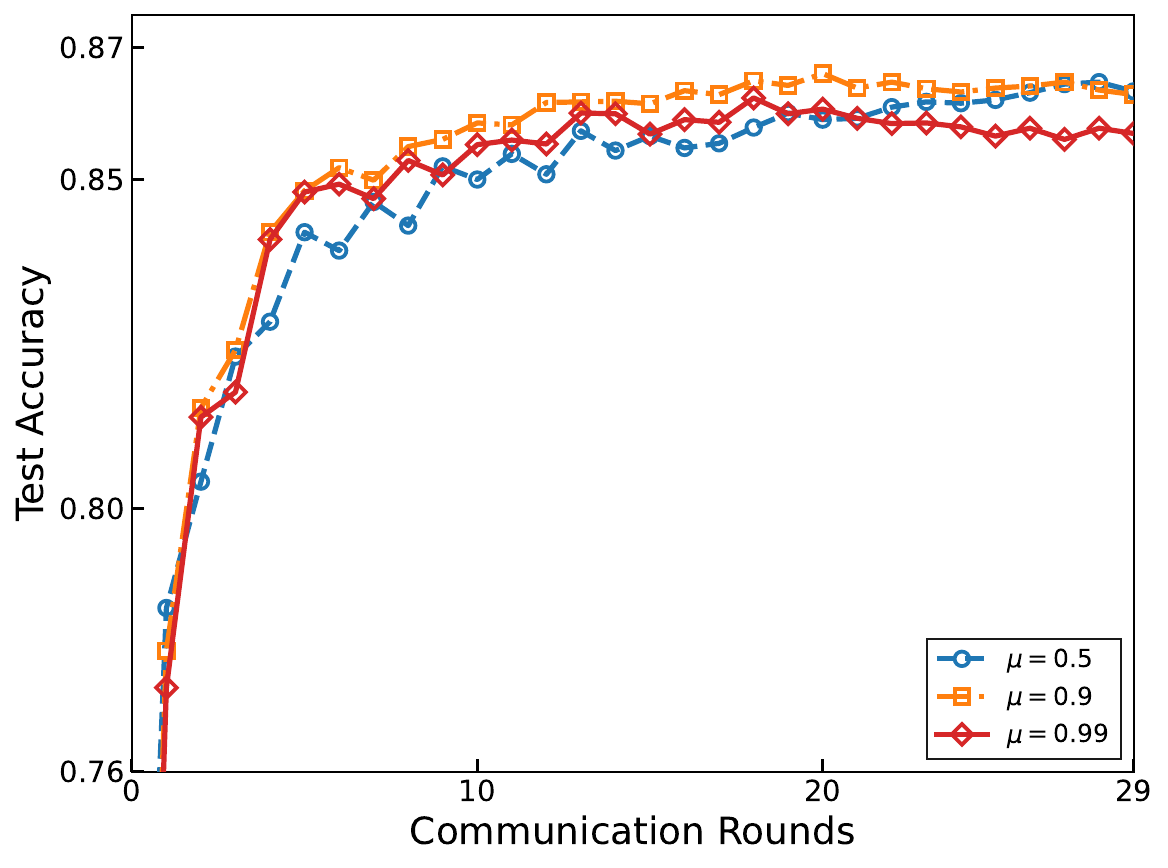}
        \caption{Momentum coefficient $\mu$}
        \label{fig:sens_mu}
    \end{subfigure}
    \hfill
    \begin{subfigure}[b]{0.48\linewidth}
        \centering
        \includegraphics[width=\linewidth]{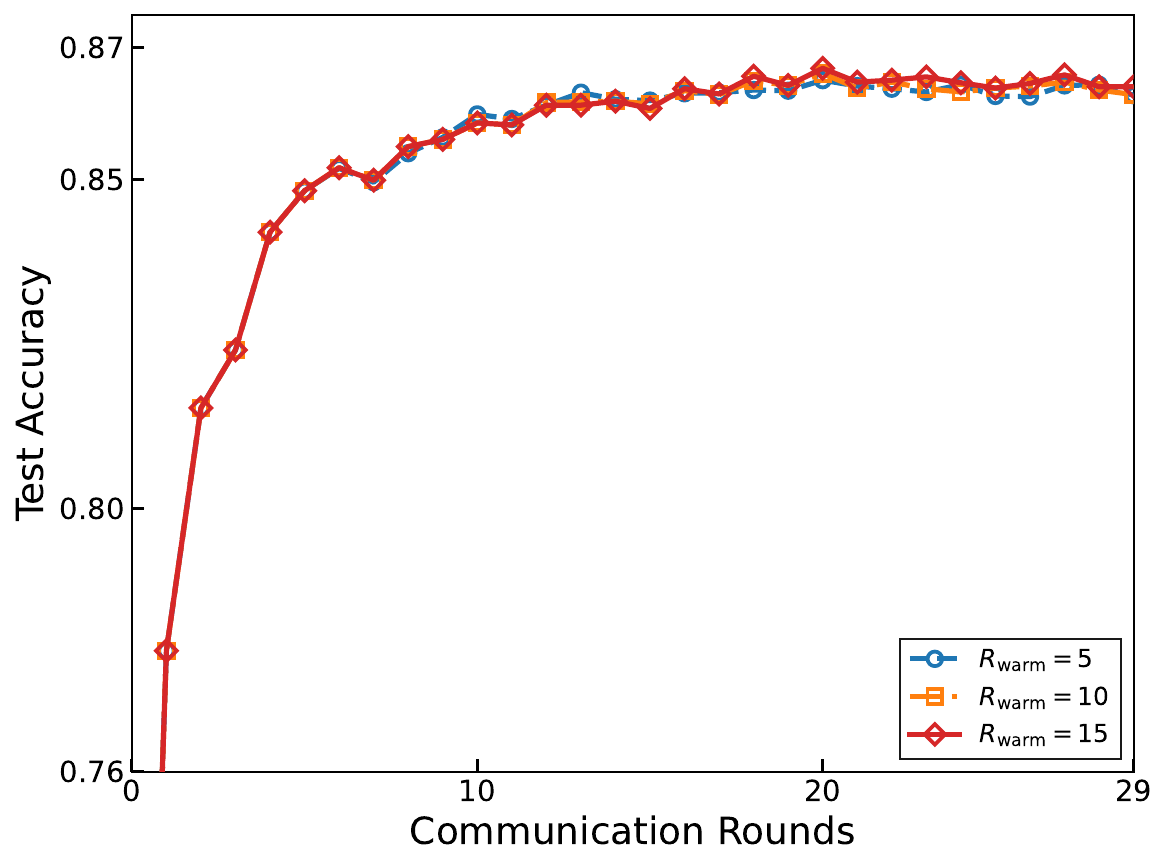}
        \caption{Warmup rounds $R_{\mathrm{warm}}$}
        \label{fig:sens_warmup}
    \end{subfigure}
    \caption{Sensitivity to training hyperparameters on
    Fashion-MNIST ($\alpha=0.1$): momentum coefficient $\mu$ and warm-up rounds
    $R_{\mathrm{warm}}$. Sensitivity to the projection cap $k_{\mathrm{proj}}$ is reported in Appendix~\ref{app:comm_cost}.}
    \label{fig:sensitivity}
\end{figure}

\textbf{Hyperparameter Sensitivity}\quad
Figure~\ref{fig:sensitivity} examines the sensitivity of SPARK
to the momentum coefficient $\mu$ and
warm-up rounds $R_{\mathrm{warm}}$ on Fashion-MNIST with
$\alpha=0.1$. We vary
$\mu \in \{0.5, 0.9, 0.99\}$ and
$R_{\mathrm{warm}} \in \{5, 10, 15\}$. A smaller momentum slows
convergence, while an overly large momentum amplifies the
approximation noise and slightly degrades accuracy. We therefore
adopt $\mu{=}0.9$ as the best trade-off between convergence speed and final
accuracy. SPARK is robust to the warm-up schedule, with all tested
values of $R_{\mathrm{warm}}$ converging at a comparable rate and
reaching a comparable accuracy, so it can be tuned flexibly.
Sensitivity to the projection cap $k_{\mathrm{proj}}$, which governs the
communication-compression trade-off, is reported in
Appendix~\ref{app:comm_cost}.

\begin{figure}[t]
\centering
\begin{subfigure}[b]{0.48\linewidth}
    \centering
    \includegraphics[width=\linewidth]{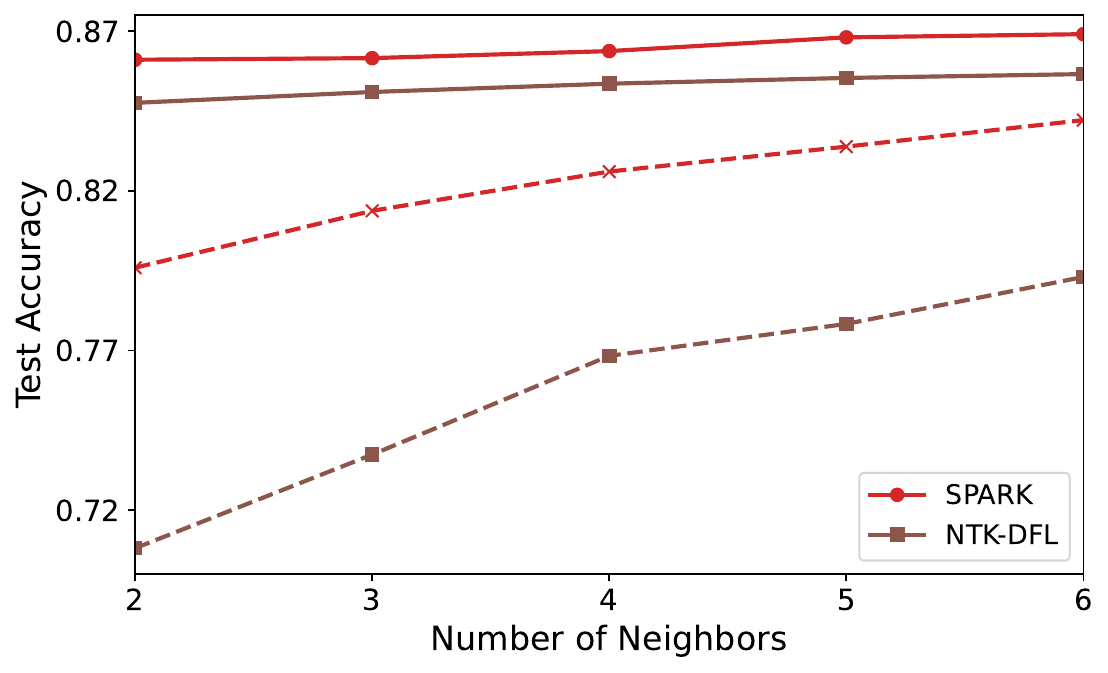}
    \subcaption{Model accuracy vs. neighbors}
    \label{fig:agg_vs_client}
\end{subfigure}
\hfill
\begin{subfigure}[b]{0.48\linewidth}
    \centering
    \includegraphics[width=\linewidth]{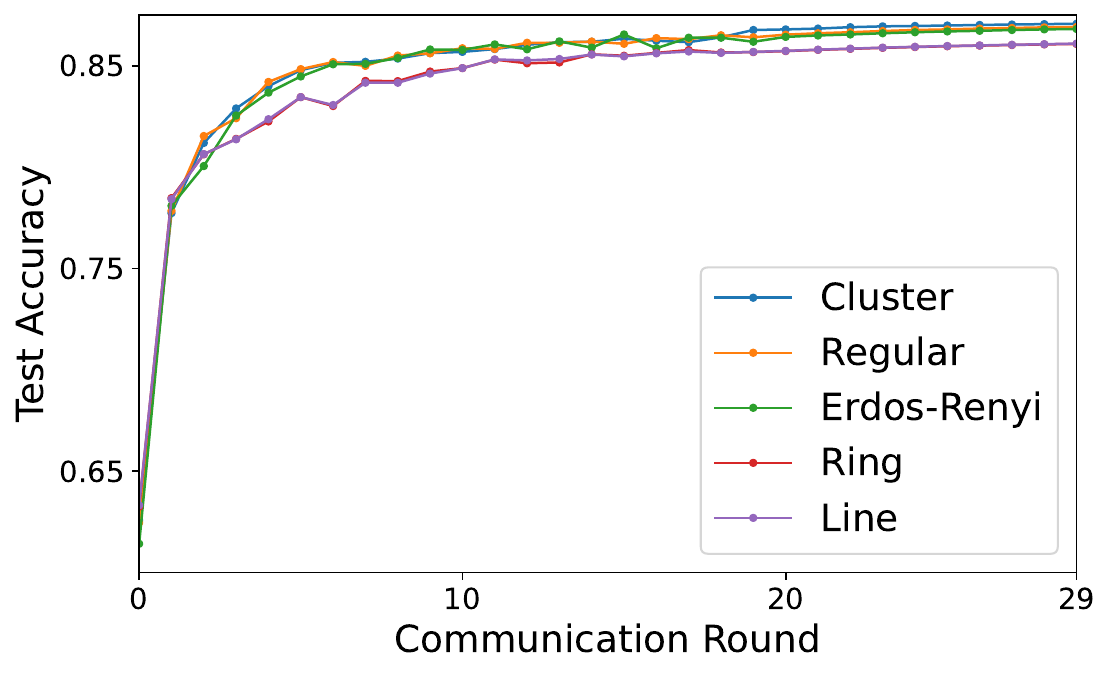}
    \subcaption{Convergence on different topologies}
    \label{fig:topology_convergence}
\end{subfigure}
\caption{
(a) Aggregated (solid) and individual client accuracy (dashed) 
vs.\ neighbor count on Fashion-MNIST ($\alpha=0.1$). 
(b) Convergence across five topology types. Cluster, regular, and
Erd\H{o}s--R\'enyi use $\kappa=5$, while ring and line use degree two.
}
\label{fig:robustness_analysis}
\vspace{-2mm}
\end{figure}

\textbf{Robustness to Network Topology}\quad 
Figure~\ref{fig:robustness_analysis} examines robustness
from two perspectives. Figure~\ref{fig:agg_vs_client} compares 
aggregated and individual client accuracy across neighbor counts. 
At $\kappa=5$, SPARK reduces the gap between aggregated 
and individual client accuracy by 4.3 percentage points compared with 
NTK-DFL, while improving aggregated accuracy by 1.3 percentage points. 
Figure~\ref{fig:topology_convergence} examines convergence across five
topology classes including cluster, regular, and Erd\H{o}s--R\'enyi graphs with
average degree $\kappa=5$, together with sparser ring and line graphs of
degree two. SPARK achieves consistent convergence trajectories across
different graph structures, indicating robustness to
topology variations.

\begin{figure*}[!t]
    \centering
    \begin{minipage}[t]{0.47\textwidth}
        \centering
        \includegraphics[width=\linewidth]{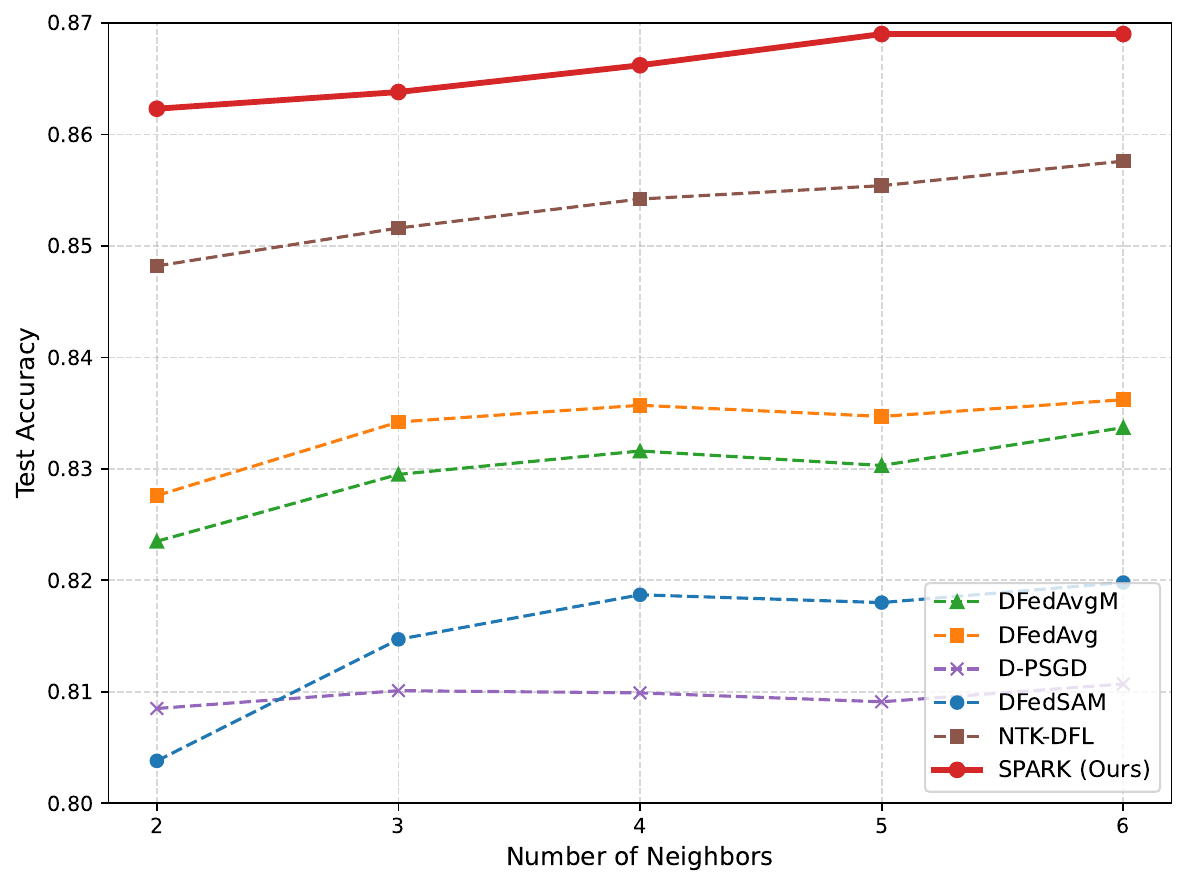}
        \caption{
        Test accuracy across network sparsity levels on Fashion-MNIST 
        ($\alpha=0.1$). Neighbor count $\kappa \in \{2, 3, 4, 5, 6\}$.
        }
        \label{fig:network_sparsity}
    \end{minipage}
    \hfill
    \begin{minipage}[t]{0.47\textwidth}
        \centering
        \includegraphics[width=\linewidth]{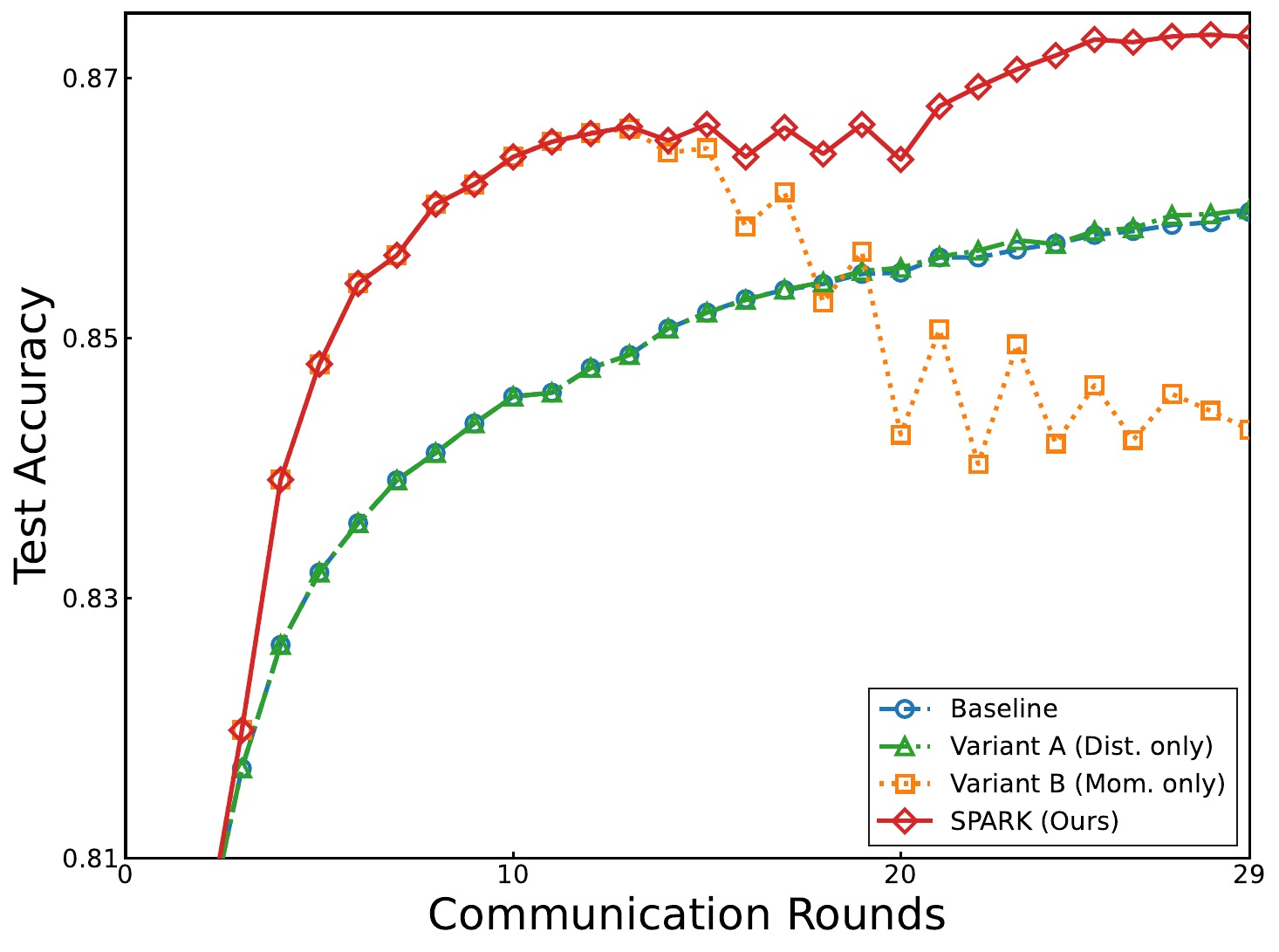}
        \caption{Convergence comparison of ablation variants under IID setting on Fashion-MNIST.}
        \label{fig:ablation_convergence}
    \end{minipage}
    \vspace{-2mm}
\end{figure*}

Figure~\ref{fig:network_sparsity} 
illustrates SPARK's performance across varying network connectivity levels with neighbor counts $\kappa \in \{2, 3, 4, 5, 6\}$ on Fashion-MNIST with $\alpha=0.1$. SPARK consistently improves over NTK-DFL and traditional decentralized baselines across all connectivity levels, showing robustness to network sparsity. The performance gap between SPARK and baselines remains stable as network connectivity varies, with SPARK maintaining accuracy between 86.3\% and 86.9\% across different $\kappa$ values. 

\textbf{Performance Under Data Heterogeneity}\quad
Table~\ref{tab:heterogeneity_robustness} presents final test accuracy 
under varying heterogeneity controlled by Dirichlet parameter 
$\alpha \in \{0.1, 0.3, 0.5\}$. SPARK consistently 
outperforms all baselines, and the improvement over NTK-DFL increases 
under more heterogeneous settings.

\begin{table*}[!t]
    \scriptsize
    \centering
    \begin{minipage}[t]{0.48\textwidth}
        \vspace{0pt}
        \centering
        \caption{Paired $t$-test of SPARK over NTK-DFL on Fashion-MNIST.}
        \label{tab:significance}
        \vspace{2mm}
        \resizebox{\linewidth}{!}{
        \begin{tabular}{lcccc}
            \toprule
            \textbf{Setting} & \textbf{Gain} (\%) & \textbf{95\% CI} & \textbf{$p$} & \textbf{$d_z$} \\
            \midrule
            IID          & $1.35$ & $[0.87,\,1.82]$ & $0.0066$ & $7.08$ \\
            $\alpha=0.1$ & $1.36$ & $[1.24,\,1.49]$ & $<0.001$ & $27.09$ \\
            $\alpha=0.3$ & $1.06$ & $[0.90,\,1.22]$ & $0.0012$ & $16.44$ \\
            $\alpha=0.5$ & $0.78$ & $[0.53,\,1.03]$ & $0.0056$ & $7.72$ \\
            \bottomrule
        \end{tabular}
        }
    \end{minipage}
    \hfill
    \begin{minipage}[t]{0.48\textwidth}
        \vspace{0pt}
        \centering
        \caption{Ablation study on SPARK components under IID setting on Fashion-MNIST.}
        \label{tab:ablation_components}
        \vspace{2mm}
        \resizebox{\linewidth}{!}{
        \begin{tabular}{cc|cccc}
            \toprule
            \textbf{Dist.} & \textbf{Mom.}
            & \textbf{Rounds}
            & \textbf{Agg. Acc.} & \textbf{Client Acc.} & \textbf{Time} \\
            & & to 85\% & (\%) & (\%) & (min) \\
            \midrule
            \ding{55} & \ding{55} & $14$ & $85.97 \pm 0.20$ & $85.24 \pm 0.07$ & $123.0$ \\
            \ding{51} & \ding{55} & $14$ & $85.99 \pm 0.09$ & $85.55 \pm 0.10$ & $121.4$ \\
            \ding{55} & \ding{51} & $6$  & $84.29 \pm 0.79$ & $86.16 \pm 0.13$ & $133.6$ \\
            \ding{51} & \ding{51} & $\mathbf{6}$ & $\mathbf{87.32 \pm 0.08}$ & $\mathbf{86.50 \pm 0.03}$ & $\mathbf{134.9}$ \\
            \bottomrule
        \end{tabular}
        }
    \end{minipage}
\end{table*}

\subsection{Ablation Study}
To validate the contribution of each core component, we evaluate all
combinations of stage-wise annealed distillation and momentum
acceleration under the IID setting in
Table~\ref{tab:ablation_components}. We refer to distillation only as
Variant A and momentum only as Variant B. These ablations isolate the
two core components. Random projection is an orthogonal
communication-reduction strategy and is analyzed separately in
Appendix~\ref{app:comm_cost}.

Table~\ref{tab:ablation_components} shows that Variant A, which
applies distillation only, improves both client generalization and aggregated
accuracy over the baseline. Variant B reduces the rounds to 85\% accuracy from 14 to 6 and achieves higher client accuracy. However,
Figure~\ref{fig:ablation_convergence} reveals that Variant B exhibits instability after round 15, with accuracy declining.
This decline results from noise amplified by momentum,
which is stabilized in complete SPARK through annealed distillation.

Complete SPARK combines distillation with momentum, where
distillation stabilizes momentum-driven training and lets SPARK keep its rapid
convergence while reaching higher aggregated accuracy.
Table~\ref{tab:ablation_components} shows that complete SPARK attains the best final accuracy among the ablation variants.
Although SPARK incurs slightly higher per-round computation than NTK-DFL due to momentum and distillation, it reaches 85\% accuracy in 6 rounds versus 14 for NTK-DFL, resulting in lower total time to target accuracy.

\section{Conclusion and Future Work}

In this paper, we have introduced SPARK, a decentralized federated learning
framework that enables stable momentum acceleration of NTK updates under
statistical heterogeneity. By co-designing stage-wise annealed distillation and
momentum, SPARK converges up to about 3$\times$ faster than NTK-based baselines
with higher accuracy, and reduces the total communication to reach a target
accuracy by up to about 70\% through fewer rounds. Random projection further
reduces per-round Jacobian communication and is analyzed as an optional
compression strategy. Validation across multiple datasets, network topologies,
and heterogeneity levels supports the robustness of the approach.

There are several promising directions for future work. For instance, extending
SPARK to deeper architectures such as CNNs and ResNets~\cite{resnet}, and evaluating it on
larger-scale image benchmarks such as CIFAR-100~\cite{cifar100}, would further test its
generality. Scaling to larger decentralized networks and measuring end-to-end
latency and energy on real IoT hardware under protocols such as
LoRaWAN, together with reducing the computational overhead
of Jacobian construction, are also worthwhile. Additionally, studying robustness
under Byzantine clients and dynamic participation, and extending the framework
beyond image classification to NLP and time-series domains, would broaden its
scope. Lastly, it is important to incorporate mechanisms that preserve the
privacy of the exchanged logits in sensitive applications.

\appendix

\section{My appendix}        
\label{app:theory}

\subsection{Hyperparameter Settings}
\label{app:hyperparams}

For the main SPARK experiments, we use a learning rate of $0.01$, full-Jacobian
updates with $\mathbf{P}_\ell=\mathbf{I}$, and Nesterov momentum $\mu=0.9$. Each
device draws a local batch of $200$ samples and evaluates its Jacobian in chunks
of $50$ to bound memory. The soft-label regularization is annealed over
communication rounds, so that hard labels dominate early training and soft
targets are phased in gradually. The evolution timestep grid $\mathcal{T}$ serves
as an implicit within-round stopping criterion for kernel evolution rather than
validation-based early stopping. We keep no held-out validation partition and
apply no round-level early stopping in the reported comparisons. A cyclic
learning-rate boost is applied at preset rounds to help escape local optima. For
the projection-enabled experiments, the projection cap is set per figure. For
every baseline, we use the hyperparameters reported in the original papers.
We report Fashion-MNIST test accuracy at communication round $30$ under
$\alpha=0.1$, which is the primary setting used in our comparisons. All
methods use a single gossip step per round, so their communication structure is
identical. Random projection is central to our compression. A random sketch
approximately preserves inner products, and thus the kernel geometry that drives
NTK evolution, which lets SPARK compress its Jacobians aggressively with little
loss of accuracy (Appendix~\ref{app:comm_cost}). Gradient-based baselines lack
this property, so we compress them with quantization alone, following
Sun et al.~\cite{dfedavg}, for a fair comparison.

\subsection{Communication Cost}
\label{app:comm_cost}

Compared with weight-based methods that communicate model parameters each round,
NTK-based updates transmit Jacobian matrices and therefore incur a higher
per-round communication cost. As introduced in Section~\ref{subsec:proposed},
SPARK reduces this cost through random projection, a strategy that is orthogonal
to the convergence behavior studied in the main text. The main results
therefore use no projection, and we report the communication-accuracy trade-off
separately here. In the uncompressed setting, each device transmits a Jacobian
of shape $N_i \times C \times d_{\mathrm{full}}$, where $N_i$ is the number of
local samples, $C$ the number of classes, and $d_{\mathrm{full}}$ the full
parameter dimension. SPARK reduces this cost through four orthogonal mechanisms,
namely projection, subsampling, sparsification, and quantization, combined with
cluster-level Jacobian reuse that makes the per-round cost independent of
neighborhood size.

\begin{figure}[!t]
    \centering
    \begin{minipage}[b]{0.40\textwidth}
        \centering
        \includegraphics[width=\linewidth]{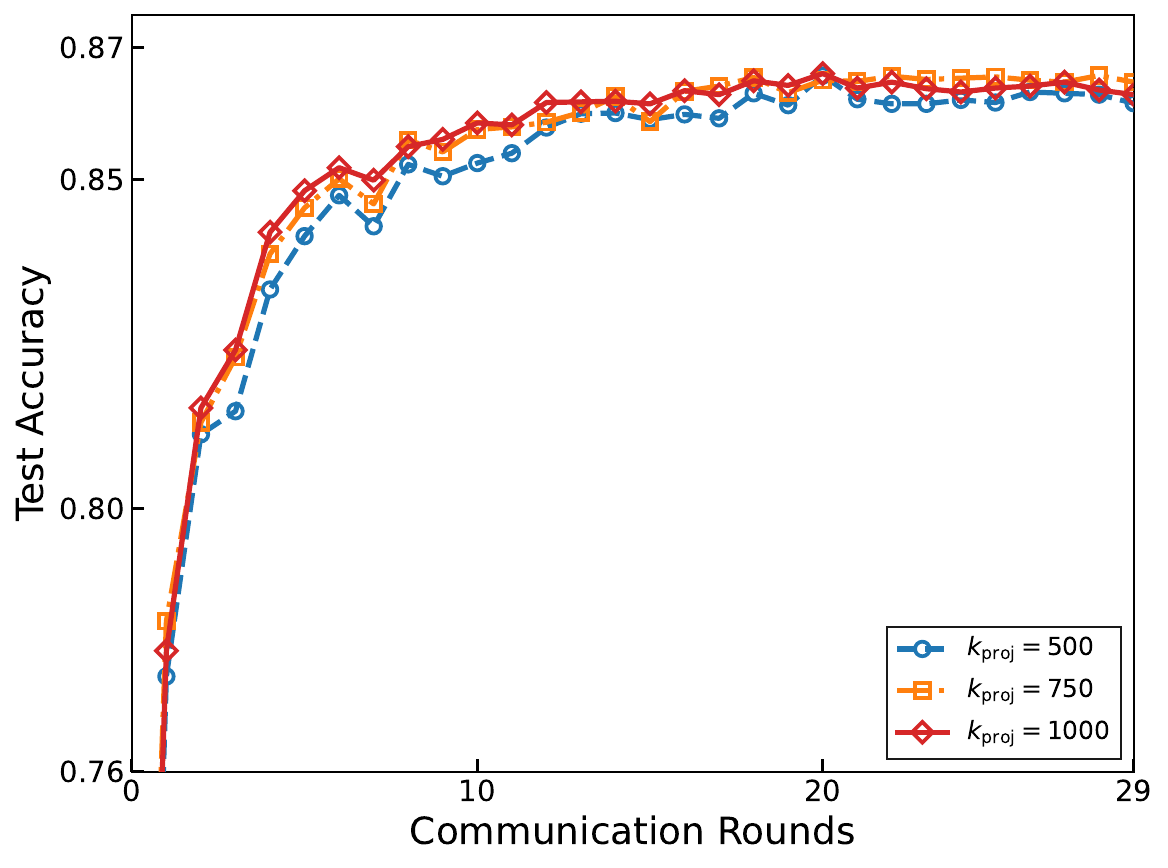}
        \caption{Test accuracy across communication rounds for
        axis-wise projection at caps $k_{\mathrm{proj}}$ of $500$, $750$, and
        $1000$ on Fashion-MNIST at $\alpha=0.1$. A smaller cap trades little
        accuracy for larger compression.}
        \label{fig:sens_k}
    \end{minipage}
    \hfill
    \begin{minipage}[b]{0.49\textwidth}
        \centering
        \includegraphics[width=\linewidth]{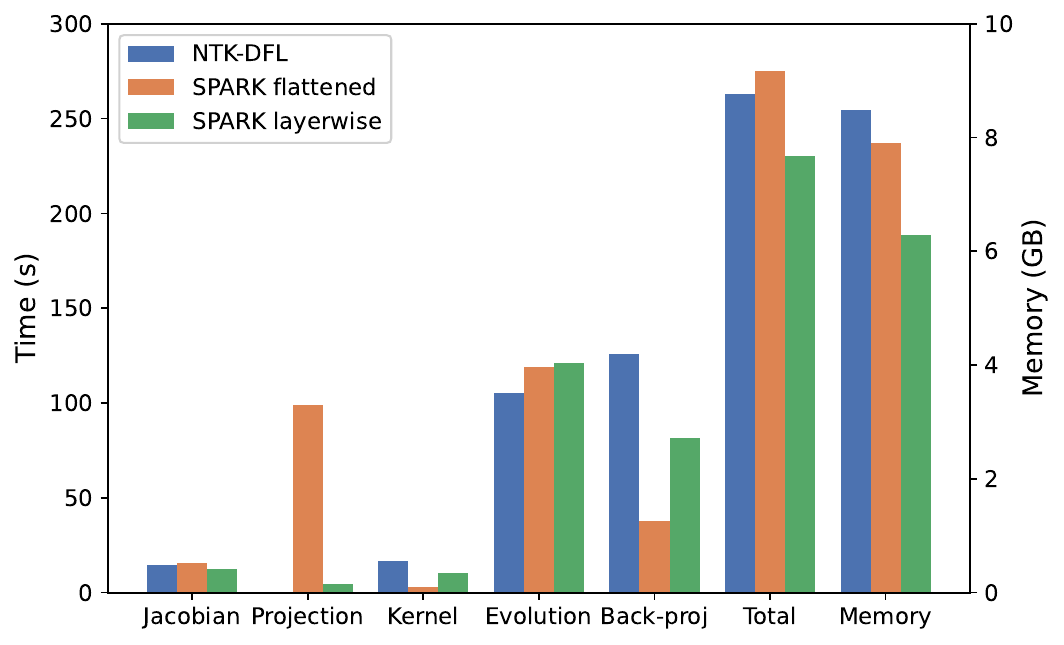}
        \caption{Per-round wall-clock time (left axis) and peak GPU memory (right axis) on Fashion-MNIST ($\alpha=0.1$). Projection reduces kernel construction, back-projection, and peak memory, keeping SPARK at or below NTK-DFL.}
        \label{fig:walltime_breakdown}
    \end{minipage}
\end{figure}

\textbf{Random projection}\quad
SPARK applies a shared deterministic Gaussian sketch, so no projection state is
communicated. The axis-wise scheme follows NTK-DFL and sketches each layer along
its last parameter axis, mapping $d_\ell^{\mathrm{in}}$ to
$k_\ell=\min(d_\ell^{\mathrm{in}},k_{\mathrm{proj}})$ while retaining the $o_\ell$ output rows,
which gives $d_{\mathrm{proj}}=\sum_\ell o_\ell k_\ell$. Figure~\ref{fig:sens_k}
shows that accuracy stays robust across the tested caps. The flattened scheme
instead vectorizes the entire parameter tensor before sketching, which gives
$d_{\mathrm{proj}}=\sum_\ell \min(o_\ell d_\ell^{\mathrm{in}},k_{\mathrm{proj}})$. By exposing
the full cross-parameter structure of a layer to one projection rather than
sketching a single axis, flattening reaches a markedly higher reduction at the
same cap, and we recommend it for bandwidth-constrained settings.

\textbf{Computational overhead}\quad
Projection added no net computational cost to SPARK. We measured the per-round
wall-clock time and peak GPU memory of each variant under the full local batch,
with the variants differing only in projection. Figure~\ref{fig:walltime_breakdown}
reports the time broken down into Jacobian computation, projection, kernel
construction, prediction evolution, and back-projection, together with the peak
memory on the right axis. Projection reduced both kernel construction and
back-projection, because these operations ran in the lower-dimensional compressed
space, and the only added term was the projection itself, which stayed small for
the layerwise scheme and was largely offset by these savings for the flattened
scheme. As a result, the per-round time of SPARK stayed comparable to NTK-DFL for
the flattened scheme and fell below it for the layerwise scheme, and the peak
memory decreased under both schemes.

\textbf{Data subsampling}\quad
Each device evolves its Jacobian on a $1/m$ fraction of its local data, which
contributes a further factor of $m$ on top of projection.

\begin{figure}[!t]
    \centering
    \begin{minipage}[b]{0.45\textwidth}
        \centering
        \includegraphics[width=\linewidth]{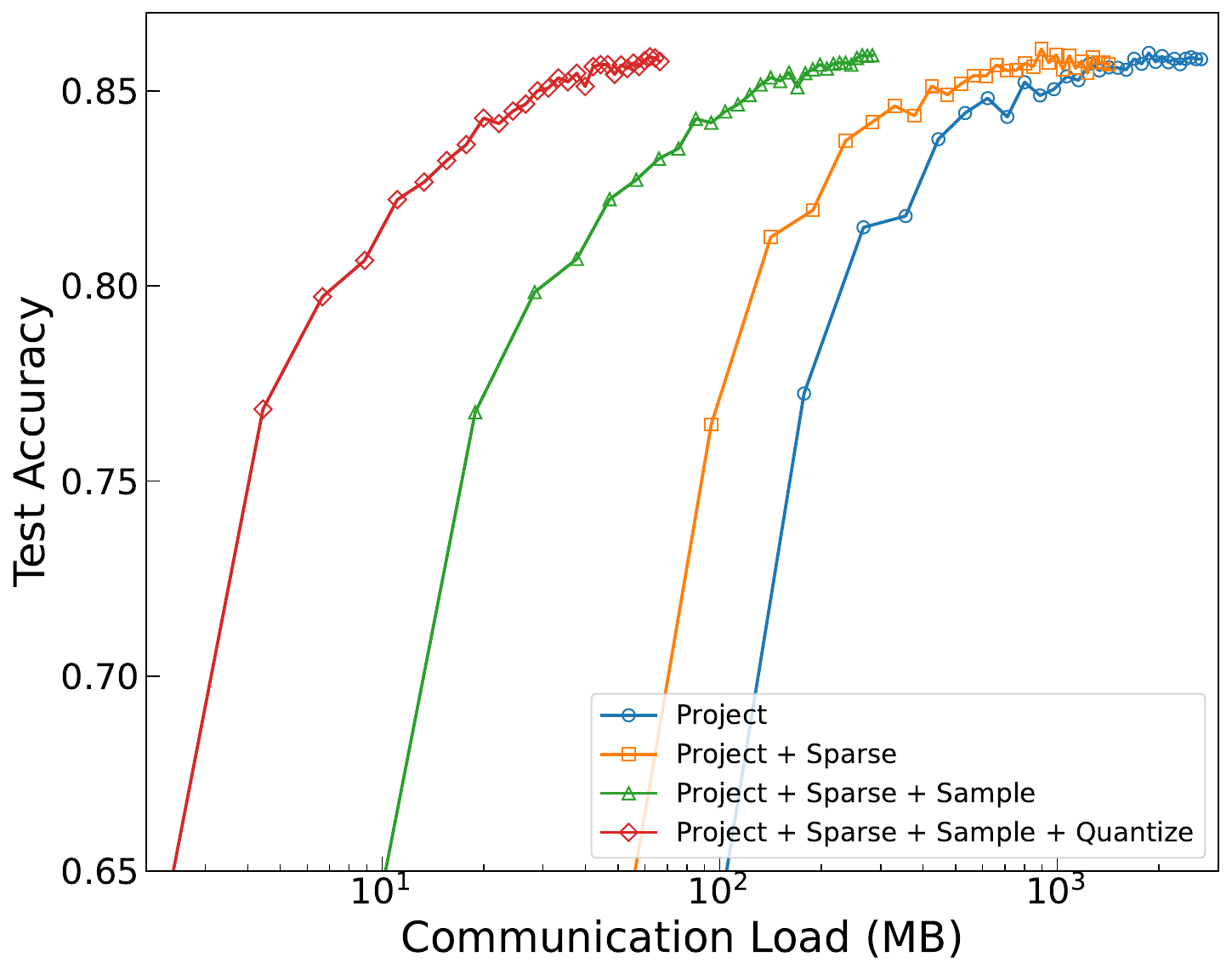}
        \caption{Progressive compression on Fashion-MNIST
        ($\alpha=0.1$). Projection, sparsification, sampling, and quantization
        compound to lower communication load at comparable accuracy.}
        \label{fig:progressive_compression}
    \end{minipage}
    \hfill
    \begin{minipage}[b]{0.49\textwidth}
        \centering
        \includegraphics[width=\linewidth]{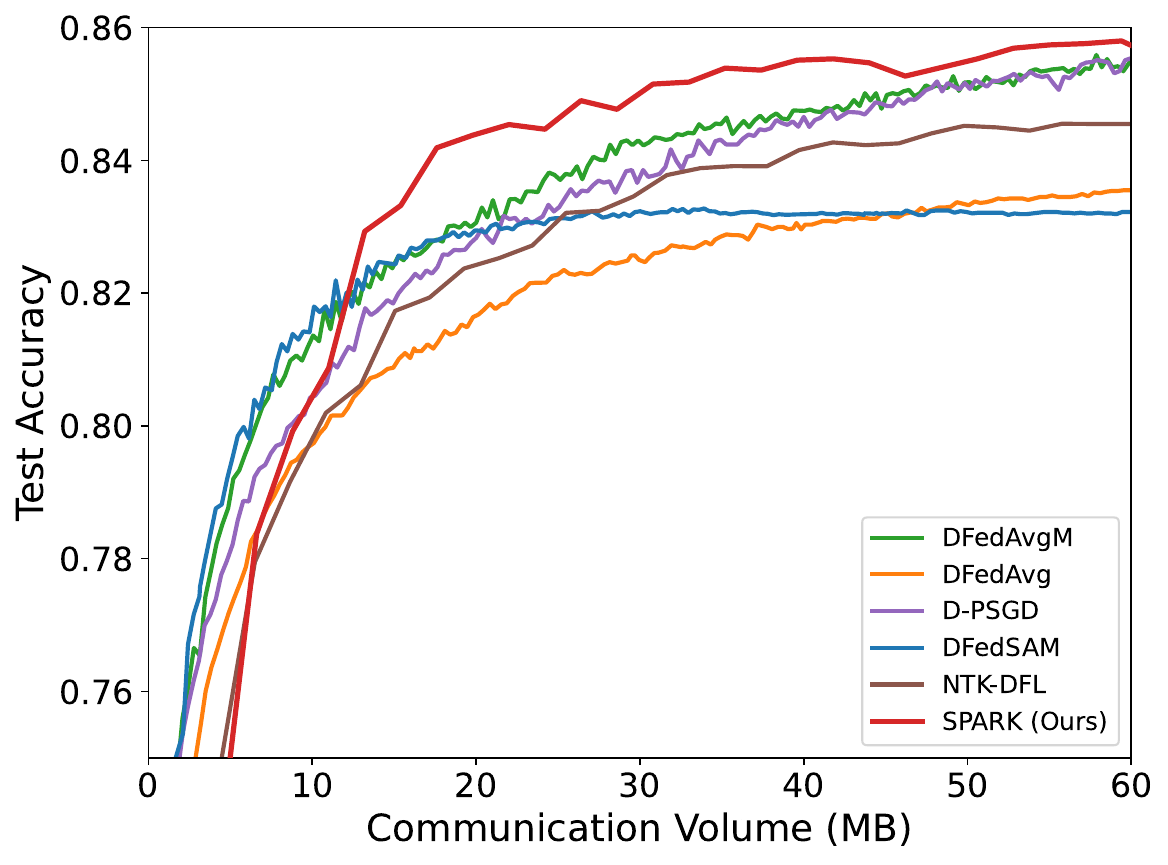}
        \caption{Accuracy versus communication volume on
        Fashion-MNIST ($\alpha=0.1$). SPARK reaches a target accuracy at lower
        volume and higher final accuracy than NTK-DFL and the baselines.}
        \label{fig:comm_volume_comparison}
    \end{minipage}
\end{figure}

\textbf{Jacobian compression}\quad
We retain the largest fraction $\rho_{\mathrm{sp}}$ of the projected entries, quantize the
survivors to $b$ bits, and signal the sparsity pattern with a compact bitmap.
These mechanisms are mutually orthogonal, so the per-package compression is the
product $(d_{\mathrm{full}}/d_{\mathrm{proj}})\,m\,
(1/\rho_{\mathrm{sp}})\,(32/b)$.
Figure~\ref{fig:progressive_compression} shows the relative communication load
for different combinations of these techniques, with a sparsification of $0.5$,
quantization to $6$ bits, a sampling of $m=5$, and a flattened projection to
$k_{\mathrm{proj}}=10000$ for the fully optimized curve.

\textbf{Clustered topology}\quad
Within a cluster, all devices share the same aggregated weight, so each device
computes a single Jacobian that is reused by all of its neighbors. The number
of Jacobians exchanged per round therefore equals the number of devices and
becomes independent of neighborhood size. As shown in
Figure~\ref{fig:comm_volume_comparison}, the full stack lets SPARK reach the
target accuracy first and attain the highest final accuracy among all methods,
which confirms that the compression preserves the convergence benefits of
NTK-based updates.

\subsection{Additional Results }
\label{app:cifar10}

\begin{figure}[!t]
    \centering
    \includegraphics[width=0.6\linewidth]{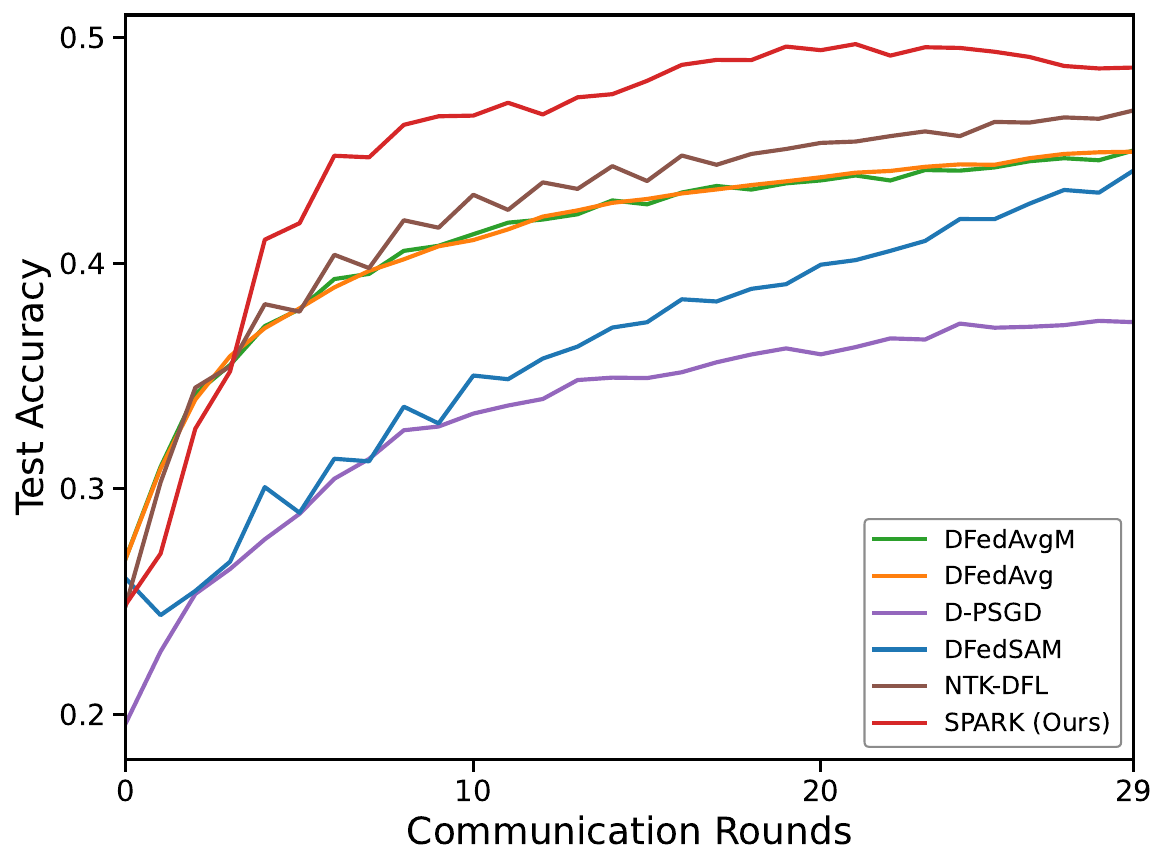}
    \caption{Test accuracy across communication rounds on non-IID
    CIFAR-10 ($\alpha=0.1$) with a two-layer MLP. SPARK converges faster and
    reaches a higher final accuracy than NTK-DFL and the decentralized
    baselines.}
    \label{fig:cifar10_convergence}
\end{figure}

To test whether the convergence advantage of SPARK extends to natural images, we
repeat the decentralized comparison on the non-IID CIFAR-10 dataset under the
same protocol as the main experiments, with a two-layer MLP and $\alpha=0.1$. As
shown in Figure~\ref{fig:cifar10_convergence}, SPARK again converges faster than
all baselines and reaches the highest final accuracy. This advantage is
consistent with the Fashion-MNIST, FEMNIST, and MNIST results, which confirms
that stabilized momentum-accelerated NTK updates remain effective on a harder
dataset.

\subsection{Mathematical Analysis}
\label{app:math_analysis}

\begin{lemma}[Sketched descent direction via back-projection]
\label{lem:backproj_descent}
Given an arbitrary gradient direction $g\in\mathbb R^{d_{\mathrm{full}}}$,
let $P\in\mathbb R^{d_{\mathrm{full}}\times d_{\mathrm{proj}}}$ be the
effective random projection matrix. For a dense Gaussian sketch, its
entries are sampled from $\mathcal N(0,1/d_{\mathrm{proj}})$. For the
layer-wise construction in \cref{eq:projection_matrix}, the effective
matrix is
$P=\mathrm{blkdiag}(I_{o_1}\otimes P_1,\ldots,
I_{o_{N_{\mathrm{lay}}}}\otimes P_{N_{\mathrm{lay}}})$.
Define the compressed gradient
$\tilde g\triangleq P^\top g\in\mathbb R^{d_{\mathrm{proj}}}$ and the back-projected direction
\begin{equation}
\label{eq:app_backproj_direction}
g_P \;\triangleq\; P\tilde g \;=\; PP^\top g .
\end{equation}
Then, over the Gaussian draw of $P$,
\begin{equation}
\label{eq:app_unbiased_stmt}
\mathbb E_P[g_P] = g.
\end{equation}
Moreover, for any fixed $g\neq 0$ and $\epsilon\in(0,1)$, on the event
that $P$ satisfies the Johnson--Lindenstrauss condition
\begin{equation}
\label{eq:app_jl_condition}
(1-\epsilon)\|g\|^2
\;\le\;
\|P^\top g\|^2
\;\le\;
(1+\epsilon)\|g\|^2 ,
\end{equation}
it follows that
\begin{equation}
\label{eq:app_descent_inner}
\langle g,\,g_P\rangle
 \ge
(1-\epsilon)\|g\|^2 >0 .
\end{equation}
Consequently, $-g_P$ is a descent direction for any objective with
gradient $g$.
\end{lemma}

\begin{proof}
\noindent The compressed Jacobian satisfies $\tilde J=JP$, so the compressed-space
gradient induced by a given residual vector is $\tilde g=P^\top g$.
Applying back-projection yields
\begin{equation}
\label{eq:app_backproj_grad}
g_P = P\tilde g = PP^\top g = S_P\,g,
\end{equation}
where $S_P=PP^\top$ is the sketching operator defined in
\cref{eq:sketch_op}. Hence the back-projected update is a
rank-$d_{\mathrm{proj}}$ sketch
of the full gradient, rather than its exact recovery.

\noindent By the Gaussian normalization in \cref{eq:projection_matrix}, each
layer-wise projection satisfies
\begin{equation}
\label{eq:app_proj_expectation}
\mathbb E_P\!\left[P_\ell P_\ell^\top\right]_{ab}
=
\sum_{r=1}^{k_\ell}\mathbb E[(P_\ell)_{ar}(P_\ell)_{br}]
=
\delta_{ab}.
\end{equation}
\noindent Therefore
$\mathbb E_P[P_\ell P_\ell^\top]=I_{d_\ell^{\mathrm{in}}}$ for each
layer. Since the corresponding effective block is
$I_{o_\ell}\otimes P_\ell$, its expected Gram matrix is
$I_{o_\ell d_\ell^{\mathrm{in}}}$. The off-block entries are zero under
the block-diagonal construction. Hence
$\mathbb E_P[S_P]=\mathbb E_P[PP^\top]=I_{d_{\mathrm{full}}}$.
The same identity follows directly for the dense Gaussian sketch.

\noindent Linearity of expectation gives
$\mathbb E_P[g_P]=\mathbb E_P[S_P]\,g=g$.

\noindent The inner product between $g$ and $g_P$ satisfies
\begin{equation}
\label{eq:app_alignment}
\langle g,\,g_P\rangle
=
g^\top PP^\top g
=
\|P^\top g\|^2 .
\end{equation}
\noindent Under condition~\cref{eq:app_jl_condition} the right-hand side is at
least $(1-\epsilon)\|g\|^2$, which is strictly positive for
$\epsilon<1$. Therefore $g_P$ makes an acute angle with $g$, and
$-g_P$ is a descent direction for sufficiently small step sizes.

\noindent The approximation error of the sketch is
\begin{equation}
\label{eq:app_projection_residual}
g - g_P = (I_{d_{\mathrm{full}}} - S_P)\,g .
\end{equation}
\noindent This residual is precisely the source of the projection component
$\delta_{\mathrm{proj}}(k_{\mathrm{proj}})$ in \cref{eq:delta_decomp}. Together with
the finite-width NTK error $\delta_{\mathrm{ntk}}$ and the
graph-induced mismatch $\delta_{\mathrm{graph}}$, it constitutes the
aggregate bias $\delta_{\mathrm{total}}$ posited in
Assumption~\ref{ass:bias}.

\noindent Thus, we have completed the proof.
\end{proof}

\begin{lemma}[Virtual iterate recursion]
\label{lem:virtual_rec}
Let $\bar w^{(k)}=\frac1M\sum_{i=1}^M w_i^{(k)}$,
$\bar v^{(k)}=\frac1M\sum_{i=1}^M v_i^{(k)}$, and
$\bar\Delta^{(k)}=\frac1M\sum_{i=1}^M \Delta w_i^{(k)}$.
Under the averaged dynamics induced by \cref{eq:spark_update_main},
define the virtual sequence $\bar z^{(k)}=\bar w^{(k)}+\frac{\mu^2}{1-\mu}\bar v^{(k)}$.
Then for all $k$,
\begin{equation}
\label{eq:z_rec_app_refined}
\bar z^{(k+1)}=\bar z^{(k)}+\frac{1}{1-\mu}\bar\Delta^{(k)}.
\end{equation}
\end{lemma}

\begin{proof}
\noindent Averaging \cref{eq:spark_update_main} over IoT devices gives
\begin{equation}
\label{eq:avg_updates_app}
\bar v^{(k+1)}=\mu \bar v^{(k)}+\bar\Delta^{(k)},
\quad
\bar w^{(k+1)}=\bar w^{(k)}+\mu \bar v^{(k+1)}+\bar\Delta^{(k)}.
\end{equation}
\noindent By definition, $\bar z^{(k)}=\bar w^{(k)}+\frac{\mu^2}{1-\mu}\bar v^{(k)}$. Hence
\begin{align}
&\bar z^{(k+1)}-\bar z^{(k)} \nonumber\\
&=\bigl(\bar w^{(k+1)}-\bar w^{(k)}\bigr)
+\frac{\mu^2}{1-\mu}\bigl(\bar v^{(k+1)}-\bar v^{(k)}\bigr) \nonumber\\
&=\mu \bar v^{(k+1)}+\bar\Delta^{(k)}
+\frac{\mu^2}{1-\mu}\bigl(\mu \bar v^{(k)}+\bar\Delta^{(k)}-\bar v^{(k)}\bigr) \label{eq:z_diff_step1}\\
&=\mu \bar v^{(k+1)}+\bar\Delta^{(k)}
-\mu^2 \bar v^{(k)}+\frac{\mu^2}{1-\mu}\bar\Delta^{(k)} \nonumber\\
&=\mu\bigl(\mu \bar v^{(k)}+\bar\Delta^{(k)}\bigr)+\bar\Delta^{(k)}
-\mu^2 \bar v^{(k)}+\frac{\mu^2}{1-\mu}\bar\Delta^{(k)} \nonumber\\
&=\Bigl(\mu+1+\frac{\mu^2}{1-\mu}\Bigr)\bar\Delta^{(k)}
=\frac{1}{1-\mu}\bar\Delta^{(k)}, \nonumber
\end{align}
\noindent where we used $\bar v^{(k+1)}=\mu \bar v^{(k)}+\bar\Delta^{(k)}$ from \cref{eq:avg_updates_app}
and the identity
\(
\mu+1+\frac{\mu^2}{1-\mu}=\frac{(1-\mu)(1+\mu)+\mu^2}{1-\mu}=\frac{1}{1-\mu}.
\)
\noindent Rearranging yields \cref{eq:z_rec_app_refined}.
\end{proof}

\begin{proof}[Proof of Theorem~\ref{thm:spark}]
\noindent Let $\mathcal F_k$ be the filtration up to round $k$, and denote
$g_k \triangleq \nabla \mathcal L(\bar z^{(k)})$ and
$\bar\Delta_{\mathbb E}^{(k)}\triangleq \mathbb E[\bar\Delta^{(k)}\mid \mathcal F_k]$.

\noindent By $L$-smoothness of $\mathcal L$ (Assumption~\ref{ass:regular}) and Lemma~\ref{lem:virtual_rec},
\begin{align}
&\mathcal L(\bar z^{(k+1)}) \nonumber\\
&\le \mathcal L(\bar z^{(k)})
+ \Big\langle g_k,\ \bar z^{(k+1)}-\bar z^{(k)} \Big\rangle \nonumber\\
&\quad+ \frac{L}{2}\big\|\bar z^{(k+1)}-\bar z^{(k)}\big\|^2 \nonumber\\
&= \mathcal L(\bar z^{(k)})
+ \frac{1}{1-\mu}\langle g_k,\bar\Delta^{(k)}\rangle
+ \frac{L}{2(1-\mu)^2}\|\bar\Delta^{(k)}\|^2.
\label{eq:one_step_refined}
\end{align}
\noindent Taking conditional expectation given $\mathcal F_k$ gives
\begin{align}
&\mathbb E\!\left[\mathcal L(\bar z^{(k+1)})\mid \mathcal F_k\right] \nonumber\\
&\le
\mathcal L(\bar z^{(k)})
+ \frac{1}{1-\mu}\langle g_k,\bar\Delta_{\mathbb E}^{(k)}\rangle \nonumber\\
&\quad+ \frac{L}{2(1-\mu)^2}\mathbb E\!\left[\|\bar\Delta^{(k)}\|^2\mid \mathcal F_k\right].
\label{eq:condexp_refined}
\end{align}

\noindent Assumption~\ref{ass:bias} implies there exists $r^{(k)}$ such that
\[
\bar\Delta_{\mathbb E}^{(k)} = -\eta g_k + r^{(k)}, \quad \|r^{(k)}\|\le \eta \delta_{\mathrm{total}}.
\]
\noindent Hence,
\begin{align}
&\langle g_k,\bar\Delta_{\mathbb E}^{(k)}\rangle
= -\eta\|g_k\|^2 + \langle g_k, r^{(k)}\rangle \nonumber\\
&\le -\eta\|g_k\|^2 + \frac{\eta}{4}\|g_k\|^2 + \frac{1}{\eta}\|r^{(k)}\|^2 \nonumber\\
&\le -\frac{3\eta}{4}\|g_k\|^2 + \eta \delta_{\mathrm{total}}^2,
\label{eq:mean_bd_refined}
\end{align}
\noindent where we used Young's inequality and $\|r^{(k)}\|^2\le \eta^2\delta_{\mathrm{total}}^2$.

\noindent Decompose $\bar\Delta^{(k)}=\bar\Delta_{\mathbb E}^{(k)}+\xi^{(k)}$ where
$\xi^{(k)}\triangleq \bar\Delta^{(k)}-\bar\Delta_{\mathbb E}^{(k)}$.
By the ``uncorrelated across IoT devices'' part of Assumption~\ref{ass:regular} and \cref{eq:annealed_var},
\begin{align}
&\mathbb E\!\left[\|\xi^{(k)}\|^2\mid \mathcal F_k\right] \nonumber\\
&=
\frac{1}{M^2}\sum_{i=1}^M
\mathbb E\!\left[\left\|\Delta w_i^{(k)}-\mathbb E[\Delta w_i^{(k)}\mid\mathcal F_k]\right\|^2\Bigm|\mathcal F_k\right] \nonumber\\
&\le \frac{\eta^2\sigma^2\gamma_k}{M}.
\label{eq:xi_var_refined}
\end{align}
\noindent Moreover,
$\|\bar\Delta_{\mathbb E}^{(k)}\|
\le \eta\|g_k\|+\eta\delta_{\mathrm{total}}$,
so $\|\bar\Delta_{\mathbb E}^{(k)}\|^2\le 2\eta^2\|g_k\|^2+2\eta^2\delta_{\mathrm{total}}^2$.

\noindent Using $\mathbb E[\langle \bar\Delta_{\mathbb E}^{(k)},\xi^{(k)}\rangle\mid \mathcal F_k]=0$,
\begin{align}
&\mathbb E\!\left[\|\bar\Delta^{(k)}\|^2\mid \mathcal F_k\right] \nonumber\\
&= \|\bar\Delta_{\mathbb E}^{(k)}\|^2 + \mathbb E\!\left[\|\xi^{(k)}\|^2\mid \mathcal F_k\right] \nonumber\\
&\le 2\eta^2\|g_k\|^2 + 2\eta^2\delta_{\mathrm{total}}^2 + \frac{\eta^2\sigma^2\gamma_k}{M}.
\label{eq:second_moment_refined}
\end{align}

\noindent Substituting \cref{eq:mean_bd_refined}--\cref{eq:second_moment_refined} into \cref{eq:condexp_refined} yields
\begin{align}
&\mathbb E\!\left[\mathcal L(\bar z^{(k+1)})\mid \mathcal F_k\right] \nonumber\\
&\le
\mathcal L(\bar z^{(k)})
-\Big(\frac{3\eta}{4(1-\mu)}-\frac{L\eta^2}{(1-\mu)^2}\Big)\|g_k\|^2 \nonumber\\
&\quad+\Big(\frac{\eta}{1-\mu}+\frac{L\eta^2}{(1-\mu)^2}\Big)\delta_{\mathrm{total}}^2 \nonumber\\
&\quad+\frac{L\eta^2}{2M(1-\mu)^2}\sigma^2\gamma_k.
\label{eq:master_refined}
\end{align}
\noindent Under $\eta\le \frac{1-\mu}{4L}$ (cf. \cref{eq:stepsize_cond}), we have
$\frac{3\eta}{4(1-\mu)}-\frac{L\eta^2}{(1-\mu)^2}\ge \frac{\eta}{2(1-\mu)}$.

\noindent Taking full expectation of \cref{eq:master_refined}, summing over $k=0,\ldots,T-1$, and using telescoping,
\begin{align}
&\frac{\eta}{2(1-\mu)}\sum_{k=0}^{T-1}\mathbb E\|g_k\|^2 \nonumber\\
&\le \mathcal L(\bar z^{(0)})-\mathcal L^\star \nonumber\\
&\quad+\Big(\frac{\eta T}{1-\mu}+\frac{L\eta^2 T}{(1-\mu)^2}\Big)\delta_{\mathrm{total}}^2 \nonumber\\
&\quad+\frac{L\eta^2}{2M(1-\mu)^2}\sum_{k=0}^{T-1}\sigma^2\gamma_k.
\label{eq:telescoping_refined}
\end{align}
\noindent Divide both sides by $T$ and use $\min_{0\le k<T} a_k \le \frac1T\sum_{k=0}^{T-1}a_k$, then multiply by $\frac{2(1-\mu)}{\eta}$:
\begin{align}
&\min_{0\le k<T}\mathbb E\|\nabla\mathcal L(\bar z^{(k)})\|^2 \nonumber\\
&\le
\frac{2(1-\mu)\big(\mathcal L(\bar z^{(0)})-\mathcal L^\star\big)}{\eta T} \nonumber\\
&\quad+\frac{L\eta}{M(1-\mu)}\cdot \frac1T\sum_{k=0}^{T-1}\sigma^2\gamma_k \nonumber\\
&\quad+\Big(2+\frac{2L\eta}{1-\mu}\Big)\delta_{\mathrm{total}}^2 .
\label{eq:final_bound}
\end{align}

\noindent Recalling $\bar\gamma=\frac1T\sum_{k=0}^{T-1}\gamma_k$
and comparing \cref{eq:final_bound} term by term with the statement
\cref{eq:main_bound} identifies the two constants explicitly:
\begin{equation}
\label{eq:constants_explicit}
C_1=1,
\qquad
C_2=2+\frac{2L\eta}{1-\mu}.
\end{equation}
That is, the stochastic term
$\frac{L\eta\sigma^2}{M(1-\mu)}\bar\gamma$ carries the constant
$C_1=1$, while the bias floor $\delta_{\mathrm{total}}^2$ carries
$C_2$. Since $\eta\le\frac{1-\mu}{4L}$ by \cref{eq:stepsize_cond}, we
have $\frac{2L\eta}{1-\mu}\le\frac12$, so $2\le C_2\le\frac52$. Both
constants are absolute and independent of $T$, $M$, and $\sigma^2$,
which establishes \cref{eq:main_bound}.

\noindent Thus, we have completed the proof.
\end{proof}

\bibliographystyle{elsarticle-num}
\bibliography{SPARK_references}

\end{document}